\documentclass[12pt,a4paper]{article}

\usepackage[british]{babel}

\usepackage[a4paper,top=2cm,bottom=2cm,left=2.5cm,right=2.5cm,marginparwidth=1.75cm]{geometry}
\usepackage{lineno}
\usepackage{setspace}
\usepackage[authoryear]{natbib}
\usepackage{hyperref}

\usepackage[dvipsnames]{xcolor}
\usepackage{amsmath}
\usepackage{graphicx}
\usepackage[title]{appendix}
\usepackage{mathrsfs}
\usepackage{authblk}
\usepackage{amsfonts}
\usepackage{booktabs} 
\usepackage{caption}  
\usepackage{threeparttable} 
\usepackage{algorithm}
\usepackage{algorithmicx}
\usepackage{algpseudocode}
\usepackage{listings}
\usepackage{enumitem}
\usepackage{chngcntr}
\usepackage{booktabs}
\usepackage{lipsum}
\usepackage{subcaption}
\usepackage{authblk}
\usepackage[T1]{fontenc}    
\usepackage[utf8]{inputenc}
\usepackage{csquotes}       
\usepackage{diagbox}
\usepackage{babel}
\usepackage{multirow}
\usepackage{array}
\usepackage{caption}
\usepackage{graphicx}
\usepackage{placeins}
\usepackage{float}
\usepackage{appendix}      
\usepackage[utf8]{inputenc}
\usepackage{newunicodechar}
\newunicodechar{−}{\textminus}
\usepackage[british]{babel}

\newcommand{\bx}{\boldsymbol{x}}

\newcommand{\btheta}{\boldsymbol{\theta}}

\begin{document}
\title{Divide-and-conquer: towards generalizable amortized Bayesian inference for the drift diffusion model\\[1em]}
\author[1]{Yufei Wu$^{\dagger}$}
\author[2]{Shanqing Gao$^{\dagger,*}$}
\author[2]{Andreas Voss}
\author[1]{Francis Tuerlinckx}

\affil[1]{KU Leuven, Belgium}
\affil[2]{Heidelberg University, Germany}

\affil[$\dagger$]{These authors contributed equally to this work.}
\affil[*]{\textit{Corresponding author:} shanqing.gao@psychologie.uni-heidelberg.de}

\date{}

\maketitle

\begin{abstract}

\noindent The drift diffusion model (DDM) is a cornerstone of cognitive decision-making research. Although numerous estimation methods exist, researchers continue to seek inference approaches that are both fast and flexible across diverse study designs. Amortized Bayesian inference (ABI) can provide nearly instantaneous inference for complex stochastic models like the DDM, but neural networks trained for one study design cannot generalize to others. In this paper, we propose a divide-and-conquer framework that address this limitation. The core idea is that the DDM's independence assumption allows the full dataset to be decomposed into pairwise shards, each sharing a common structure that a single neural network can learn. Inference is performed on each shard separately and the resulting posteriors are combined via consensus MCMC to approximate the full posterior. Using simulated datasets, we evaluate the accuracy and uncertainty of this method. Our results show that the proposed divide-and-conquer approach achieves accuracy and uncertainty comparable to MCMC while reducing computational cost by several orders of magnitude. This work not only advances DDM estimation but also demonstrates a general strategy for improving the scalability and generalizability of ABI methods across diverse applications.

\end{abstract}


\section{Introduction}

The drift diffusion model \citep[DDM,][]{ratcliff1978theory} is one of the most widely used models of decision-making in psychology. A recent study identified more than 10,000 academic works that include the term “drift diffusion model” in their titles or abstracts \citep{priem2022openalex}. The DDM assumes that, when having to make a quick decision, participants accumulate evidence over time until it reaches a decision threshold, at which point a response is made. The widespread use of the DDM is driven by several factors: a parsimonious representation with psychologically meaningful parameters that can be linked to neural substrates \citep{ratcliff1978theory, myers_practical_2022, gupta_neural_2022}, and the availability of simple and efficient fitting procedures through various toolboxes \citep{voss2007fast, wagenmakers2007ez, ratcliff_estimating_2002}.

Having access to fast, or even near-instantaneous, estimation of DDM parameters would be of great importance in many situations. For example, a researcher would like to examine how decision-making parameters vary across age or other demographic variables in a very large dataset with over one million participants \citep[e.g.,][]{ProjectImplicit}. To the best of our knowledge, such data sizes exceed the capabilities of the current DDM software. As another example, a researcher may want to know the optimal design for a given research question through simulations. This involves repeatedly estimating DDMs with varying numbers of participants and design specifications. Without fast estimation routines, this is not feasible. Furthermore, instantaneous estimation would enable adaptive experimental designs \citep{myung2013tutorial}, where stimuli can be tailored on the fly based on information obtained from previous trials. These examples illustrate that fast online estimation not only improves efficiency but also expands the range of feasible research questions.

Recent development in amortized Bayesian inference \citep[ABI,][]{radev_bayesflow_2023} enables such fast inference. ABI leverages simulated training data to optimize generative neural networks that encode structural and functional knowledge about the simulation model and its parameters. The inference target can be the point estimates or posterior distribution \citep{zammit2024neural, radev_bayesflow_2023}. By recasting the costly inference task as forward passes through a trained neural network, ABI achieves nearly instantaneous (i.e., \textit{amortized}) parameter inference for new datasets \citep[]{gonccalves2020training, radev_bayesflow_2023}. We refer to the neural networks that learn a representation of the posterior distribution as a neural posterior estimator (NPE).

Once trained, NPEs perform inference instantaneously. On the other hand, training is more costly. Training a new NPE for each design is computationally expensive, often requiring substantial time and resources (e.g., GPUs). Moreover, when an NPE is trained for a specific experimental design, it does not generalize across experimental designs: the test data must be generated from the same observation model as the training data. These issues are further exacerbated for large datasets because simulating from the model becomes increasingly computationally demanding.

In this project, we address the generalization challenge of DDM estimation in ABI, validating our results throughout against traditional MCMC. This comparison serves two main purposes. First, although ABI has shown remarkable promise, it remains relatively new and not yet widely adopted in the cognitive modeling community, so benchmarking against the established MCMC approach helps build confidence in the method. Second, while the theoretical properties of neural posterior estimation are becoming better understood \citep{frazier2024statistical}, MCMC remains the gold standard for Bayesian inference with well-characterized properties, making it a natural reference for validating the accuracy and uncertainty quantification of ABI. 

Our approach exploits a fundamental property of the DDM: observations are independent across trials and conditions. This independence means that the full-data likelihood factorizes exactly into the product of likelihoods over any disjoint partition of the data. In other words, if we split the dataset into non-overlapping subsets and perform inference on each subset separately, the combined information is mathematically equivalent to performing inference on the full dataset at once, as long as every trial appears in exactly one subset. Building on this property, we partition the data into pairs of conditions and perform inference on each pair separately, then combine the resulting posteriors using consensus MCMC to recover the full posterior \citep{scott2022bayes}. Together, these two implementations demonstrate that the divide-and-conquer strategy is both flexible across inference methods and scalable across diverse experimental designs. 

\section{Drift diffusion model}

\begin{figure}[!h] 
 \centering
 \includegraphics[width=1\linewidth]{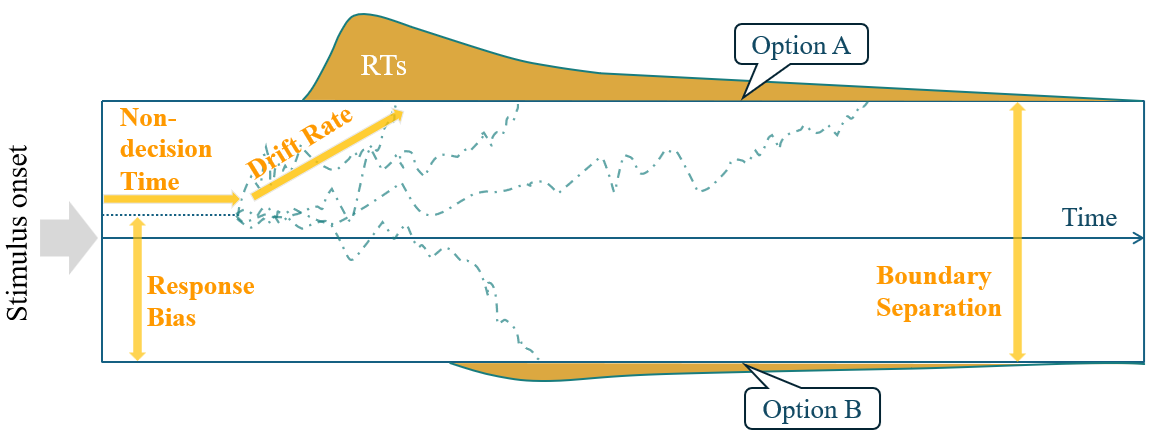}
 \caption{The schematic of the DDM. The two boundaries represent the two options in an experiment. After stimulus onset, evidence accumulates stochastically between the boundaries. The time it takes for the accumulated evidence to reach a boundary plus the non-decision time corresponds to the reaction time.} 
 \label{fig:DDM}
\end{figure}

The core assumptions of the DDM are illustrated in Figure~\ref{fig:DDM}. In a single experimental trial, a decision is made when the accumulated evidence $X_t$ reaches one of the two boundaries: 0 and $a$, and $a$ is known as boundary separation. Evidence accumulation begins at a relative starting point (also known as response bias) $z \in [0,1]$. If $z>0.5$ ($z<0.5$), then there is a bias towards the upper (lower) boundary, while if $z=0.5$, there is no bias. The drift rate $v$ represents the average rate of evidence accumulation:
\begin{equation*}
    dX_t = v \, dt + \sigma \, dW_t.
\label{eq:1dwiener}
\end{equation*} 
This equation captures the random evolution of $X$ over time $t$, with deterministic drift $v$ and $dW_t$, the differential of Brownian motion, or Wiener process \footnote{The amount of noise in the evidence accumulation process is scaled by the diffusion constant $\sigma$. In this study, a diffusion constant of $\sigma=1$ was used.}. When $X_t \geq a$ or $X_t \leq 0$, the diffusion process stops, and the decision is made. Finally, the non-decision time $T_{er}$ accounts for processes unrelated to decision formation, such as stimulus encoding and motor execution. The resulting reaction time and corresponding decision in  trial $i$ can then be registered as $x_i=(RT_i, C_i)$ ($i=1,\dots,n$).

In experiments where the DDM can be applied, each participant makes decisions under multiple conditions. Here, a \textit{condition} refers to a set of trials sharing the same DDM parameters; that is, all five DDM parameters are assumed to be identical within a condition, corresponding to a single experimental manipulation. While some parameters are shared across conditions, others may vary. For example, when studying whether stimulus difficulty affects the speed of evidence accumulation, a researcher may allow the drift rate to differ across conditions while keeping the other parameters fixed. Each condition typically contains tens to hundreds of trials to reliably estimate the DDM parameters. Importantly, the DDM assumes independence across trials and conditions, meaning the reaction time and choice on a given trial are not influenced by other trials, and the data from one condition are not influenced by data from other conditions. 


\section{Bayesian inference methods}

Bayesian inference aims to learn the posterior distribution $p(\btheta \mid \bx)$ of model parameters $\btheta$ (a vector of dimensionality $K\geq 1$) given observed data $\bx$ (also combined into a vector). The most established approach for obtaining posterior samples is MCMC sampling. MCMC generates posterior draws by constructing a Markov chain that converges to the target distribution, accepting or rejecting a proposed parameter value $\btheta^{\prime}$ based on $p(\bx \mid \btheta^{\prime})$, the likelihood of data given $\btheta^{\prime}$. While MCMC is theoretically well-understood and produces asymptotically exact posterior samples, it requires an explicit likelihood function and always performs inference from scratch for each new dataset. 

Amortized methods take a fundamentally different approach to Bayesian inference. Specifically, ABI utilizes simulated data along with their corresponding ground-truth parameters to train specialized neural networks that learn a global posterior given \textit{any} dataset compatible with the model \citep{schmitt2023detecting}. Once trained, the networks can instantly perform inference by reusing learned mappings, with the upfront training cost distributed across repeated use (i.e., \textit{amortized}). This makes ABI both likelihood-free, because researchers only need to simulate data from a given model without knowing the likelihood of the data given a parameter value, and highly scalable. For a detailed description of the ABI workflow that was used in this paper, see Appendix~\ref{app:abi}.

Despite these advantages, a key limitation of ABI is that trained neural networks (i.e., NPEs) lack generalizability. Trained networks implicitly learn both the underlying model and the structure of the training data, including the number of parameters, data size, and other design-specific features. For example, an NPE trained on choice reaction time data from an experiment with three conditions cannot be directly applied to data from an experiment with four or five conditions, because the structure and dimensionality of the inputs and outputs differ from what the network has learned to process during training.

In the context of the DDM, this limitation is particularly restrictive, as experimental designs often vary in the number and configuration of conditions. Training a separate neural network for each design is computationally expensive and undermines the efficiency gains of amortized inference. In the following section, we propose a divide-and-conquer strategy that addresses this limitation by decomposing any experimental design into condition pairs, enabling a single NPE to generalize across diverse designs without retraining.

\section{Scalable divide-and-conquer strategy}

Suppose a researcher has data from an experimental design with three conditions. The researcher wants to fit a DDM to the data with one drift rate per condition and the other parameters shared across all conditions. However, the researcher has only trained an NPE for a DDM with two conditions (varying drift rates, other parameters constant). Can the researcher use the trained NPE to fit the three-condition model without retraining?

\begin{figure}[!ht]
\centering
\includegraphics[width=1\linewidth]{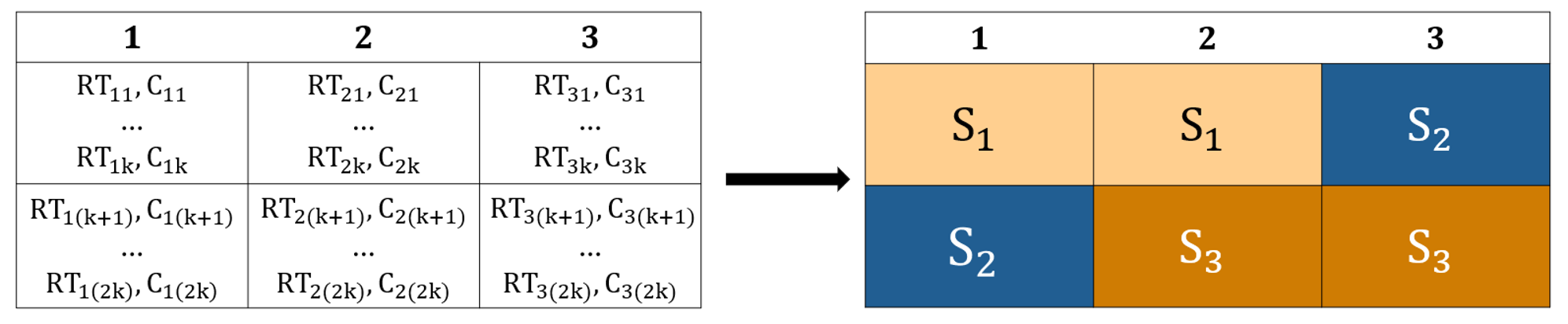}
\caption{Data partitioning example. For an experimental design with three conditions (1, 2, 3), trials within each condition are equally split into two parts, producing six subsets arranged into three shards: $S_1$, $S_2$, and ${S_3}$.}
\label{fig:grouping}
\end{figure}

To discuss the problem further, let us first define some notation. Assuming three conditions, the full model underlying the response on trial $i$ in condition $k$ is: $x_{ik}=(RT_{ik},C_{ik}) \sim \text{Wiener}(v_k, a, z, T_{er})$ with $k = 1, 2, 3$. The data of a single participant can be visualized in the wide format (Figure~\ref{fig:grouping}, left panel). The challenge is that a neural network trained on this three-condition design cannot be directly applied to data from a different number of conditions. We therefore ask: Can we decompose this dataset into smaller subsets that share a common structure, allowing a single network trained for a 2-condition model to handle this design?

One approach is to split the trials within each condition into two equal parts. This yields six subsets that can be arranged into three pairwise combinations: $S_1, S_2$ and $S_3$ (see Figure~\ref{fig:grouping}, right panel). Each pair shares the same model structure: two conditions, two drift rates, and a common set of remaining parameters, allowing a single pairwise NPE to be applied across all combinations regardless of the original experimental design.

While this decomposition may seem intuitive, it is important to establish that fitting pairwise models is not an approximation of the full model fitting but yields results exactly equivalent to fitting the full model. To see this, recall that we are interested in learning the full posterior

$$
p(\boldsymbol{\theta} \mid \bx) \propto  p(\bx|\boldsymbol{\theta}) p(\boldsymbol{\theta}),
$$
where $\bx$ refers to all trials. 

We partition the data into $Q$ disjoint shards $S_1, \dots , S_Q$, so that $\bx = (S_1, \dots , S_Q)$. Because the DDM assumes that all trials are independent given $\btheta$, and because the partition is disjoint and exhaustive (every trial appears in exactly one shard), the full-data likelihood factorizes exactly as the product of the shard likelihoods:

$$
p(\bx \mid \btheta) = \prod_{q=1}^Q p(S_q \mid \btheta).
$$

Consequently, the full posterior can be written as

$$
p(\btheta \mid \bx) \propto \prod_{q=1}^Q p(S_q \mid \btheta) p(\btheta).
$$

This shows that analyzing the shards separately preserves the full information in $\bx$, provided no trial is reused across shards. In our scenario, directly computing $p(\btheta \mid \bx)$ is challenging since a single NPE trained on pairwise data cannot directly process the full dataset. We therefore analyze the posteriors corresponding to the separate groups:
\begin{equation} \label{eq:shard_eq}
p(\btheta \mid S_q) \propto p(S_q \mid \btheta) p(\btheta)
\end{equation}
and then combine the information from the various posteriors $p(\btheta \mid S_1), \dots, p(\btheta \mid S_Q)$. 

Suppose the researcher has obtained posterior draws from each shard following the partitioning in Figure~\ref{fig:grouping}. Since the NPE is trained by assuming the ground-truth parameter comes from $p(\btheta)$, these draws come from $p(\btheta \mid S_q) \propto p(S_q \mid \btheta) p(\btheta)$ with $q = 1,\dots,Q$. We now require a method to combine them into an approximation of the full posterior.

To motivate how these draws should be combined, note that under the Bernstein-von Mises theorem, each shard posterior is approximately Gaussian for sufficiently large samples, concentrating around the true parameter value \citep{le2012asymptotic}. A useful property of Gaussian densities is that their product is itself proportional to a Gaussian. Specifically, if each shard posterior is approximated by $\mathcal{N}(\hat\mu_q, \hat\Sigma_q)$, then their product is proportional to Gaussian with precision and mean
\begin{align*}
\hat{\mu} &= \left( \sum_{q = 1}^Q\hat \Sigma_q^{-1} \right)^{-1} \left(\sum_{q=1}^Q \hat \Sigma_q^{-1}\hat \mu_q \right), \\
\hat{\Sigma} &= \left( \sum_{q = 1}^Q\hat \Sigma_q^{-1} \right)^{-1}.
\end{align*}
The combined mean is thus a precision-weighted average of the shard means: shards with higher precision contribute more to the combined estimate \citep{neiswanger2013asymptotically}.

In practice, however, the shard posteriors are available only as Monte Carlo draws rather than closed-form Gaussians. Consensus MCMC therefore applies the precision-weighted average directly to the draws \citep{scott2022bayes}. Let $\btheta_{q1},...,\btheta_{ql}$ denote the marginal posterior draws for one parameter from shard $q$, for $q = 1,\dots, Q$. The consensus posterior for draw $l$ is:
\begin{equation}
    \btheta_l = \left(\sum_{q=1}^Q W^{-1}_q \right)^{-1}\sum_{q=1}^Q W^{-1}_q\btheta_{ql},
    \label{eq:consensus}
\end{equation}
where $W_q$ is the marginal posterior covariance matrix of shard $q$ calculated from the posterior draws. The optimal scenario is to use the full covariance matrix, but we approximate this through a diagonal matrix (see \citet{scott2022bayes} for details).

When combining the $Q$ shard-specific posteriors (see Equation~\ref{eq:shard_eq}), a problem arises: the prior $p(\btheta)$ is counted $Q$ times rather than once, causing it to weigh too heavily. Therefore, a correct analysis requires distributing the prior across the $Q$ subsets
$$
p(\btheta \mid S_q) \propto p(S_q \mid \btheta) p(\btheta)^{\frac{1}{Q}}
$$
so that the total prior information is preserved. How to handle the prior overcounting in practice is described below. Reusing the prior $Q$ times distorts the consensus posterior. Indeed, when for each subset the prior $p(\theta)$ is used, the effective prior in the consensus posterior becomes $p(\theta)^Q$ rather than $p(\theta)$. This concern can be addressed via importance sampling \citep{tokdar2010importance}. Specifically, the consensus posterior draws are reweighted by $p(\theta)^{1-Q}$ to recover the correct posterior (see Appendix~\ref{app:Importance_samp} for details). When the Q is relatively small and the priors are uninformative, the distortion introduced by prior reuse is negligible.

Together, these components form the divide-and-conquer framework proposed in this paper. By partitioning any experimental design into shards of two conditions, a single pairwise NPE trained on pairwise data can be applied to any shard regardless of the original design, resolving the generalizability limitation of standard ABI. The exact likelihood factorization established above guarantees that this decomposition is not an approximation, because analyzing shards separately and combining the results via consensus MCMC with importance sampling calibration is mathematically equivalent to analyzing the full dataset at once. This strategy therefore enables fast, flexible, and theoretically exact Bayesian inference for the DDM across diverse experimental designs.

For clarity, we distinguish the following posteriors used throughout this work: a) \textit{pairwise posterior}, the posterior conditioned on each data shard (pair) $S_q$, b) \textit{consensus posterior}, the posterior draws averaged across shards via Equation~\ref{eq:consensus}, representing a consensus given the complete data, c) \textit{calibrated consensus posterior}, the consensus posterior calibrated via importance sampling to remove the impact of reusing the prior multiple times, d) \textit{full posterior}, the posterior conditioned on complete data set, serving as a benchmark to validate our approach.

\section{Simulation studies: assessing the accuracy and uncertainty of the divide-and-conquer approach}

Having established the divide-and-conquer framework, we now evaluate its performance through two simulation studies. Our central question is whether the calibrated consensus posterior can accurately approximate the full posterior obtained by fitting the complete model. To validate the approach beyond ABI, we apply the same divide-and-conquer strategy using traditional MCMC, allowing us to distinguish effects specific to the neural network from those inherent to the pairwise decomposition. We first describe the simulation settings and estimation procedures, then compare the calibrated consensus posterior obtained via the divide-and-conquer approach with the full posterior across both ABI and MCMC.

\subsection*{Setting 1}
We specified the full model
\begin{equation}
    x_{ik}=(RT_{ik},C_{ik}) \sim \text{Wiener}(v_k, a, z, T_{er}), \quad k = 1,2,3,4,
\label{eq:full1}
\end{equation}
corresponding to an experiment with four conditions. Each condition was assigned a drift rate $v_k$ (i.e., different average rates of evidence accumulation are assumed across conditions), while other parameters $(a,z, T_{er})$ were held constant across conditions. For each $k$, we simulated testing datasets with $n = 100, 200, 500$ trials per condition, and generated 100 replicates for each $n$.

\subsection*{Setting 2}
We specified the full model
\begin{equation}
    x_{ijk}=(RT_{ijk},C_{ijk})  \sim \text{Wiener}(v_j, a_k, z, T_{er}), \quad j = 1,2,3, \quad k=1,2,
\label{eq:full2}
\end{equation}
corresponding to a study with a $3 \times 2 \text{ (stimulus} \times \text{instruction)}$ design, where $j$ is the stimulus index and $k$ is the instruction index, resulting in six conditions in total. Participants were instructed to respond either as quickly as possible or as accurately as possible. Different instructions typically reveal a speed-accuracy trade-off in datasets, which can be explained by the boundary separation $a$ \citep{myers_practical_2022}: under speed instructions, the decision boundary is lower, producing faster responses at the cost of increased errors, whereas under accuracy instructions, the decision boundary is higher, yielding slower but more accurate responses. For each condition, we also simulated testing datasets with $n = 100, 200, 500$ trials per condition, and generated 100 replicates for each $n$.

\subsection*{Estimation}
We compared estimates from two approaches: pairwise estimation (the divide-and-conquer approach using a pairwise NPE applied to each shard) and full-model estimation, in both ABI and MCMC. For ABI, a pairwise NPE was trained by data simulated from an observation model $\text{Wiener}(v_k, a, z, T_{er}), k =1,2$. The full-model NPEs were trained using data simulated from the distributions suggested by Equation~\ref{eq:full1} and ~\ref{eq:full2}, respectively. Training data were simulated using the Euler-Maruyama method, where the continuous evidence accumulation process is approximated with a discrete time step $\Delta t = 0.001s$.

When simulating the training data, it is important that the trial numbers of the training data match the trial numbers in the test data. When simulating the training data for the full model NPEs, the trial number per condition $n \sim U(90, 600)$ (i.e., discrete uniform). The $n$ for the pairwise NPE is sampled from $U(30, 300)$. This is because, as Table~\ref{pairwise} suggests, when performing pairwise estimation, each test data set needs to be partitioned into six disjoint pairs; thus, the number of trials per condition is smaller.

\begin{figure}
\centering
\includegraphics[width=1\linewidth]{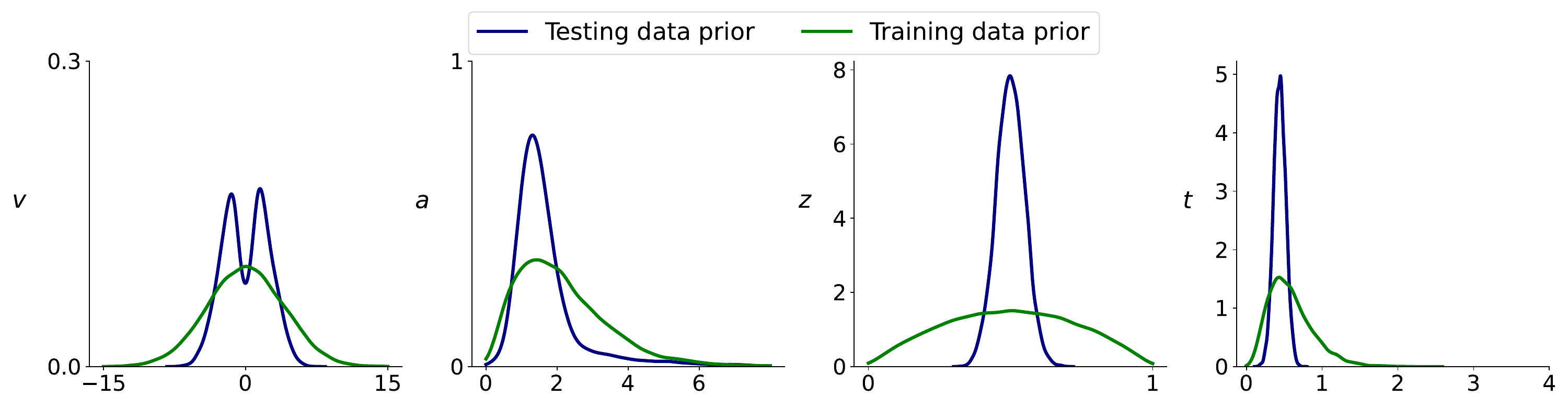}
\caption{The testing data prior and the training data prior.}\label{fig:prior}
\end{figure}

The prior for the parameters in the training data is $v \sim \mathcal{N}(0,4), a \sim \text{Gamma}(2.7, 0.8), z \sim \text{Beta}(2,2), T_{er} \sim \text{Gamma}(3.9, 0.15)$, where in the $\text{Gamma}(\alpha, \beta)$, $\alpha$ and $\beta$ denote the shape and scale parameters. The prior in the training data for the NPE is identical to the prior used in MCMC sampling, representing the initial beliefs about model parameters. The distributions of true parameters for the test data were narrower than the training data prior and were chosen based on a systematic parameter review of the DDM, see Table~\ref{tab:simprior} and Figure~\ref{fig:prior} \citep{tran2021systematic}.

The training was performed using the Python package \texttt{BayesFlow}, version 2.0.12 \citep{kuhmichel2026bayesflow}. Pairwise NPE training was run for 100 epochs (512 iterations per epoch, batch size of 64) using the Adam optimizer (initial learning rate = $5 \times 10^{-4}$, cosine decay schedule). The summary network was a Set Transformer with an output dimension of 20, and the inference network was a 4-layer multilayer perceptron (MLP) with 256 units per layer. Training (including simulation) was performed on an NVIDIA Tesla H100-SXM2-32GB GPU.

The full-model NPE in Setting 1 was trained using the same configuration, except that the summary network output dimension was increased to 64 and the inference network used 512 units per layer. The full-model NPE in Setting 2, in addition to the previous configuration changes, used 768 units per layer. After the training, all NPEs produced accurate mean estimations and well-calibrated posteriors. Results are shown in Appendix~\ref{ap:recovery}.

Next, We input the full test datasets to the full-model NPEs to obtain full posteriors (3000 posterior draws per dataset). We also input disjoint pairs of observations to the pairwise NPE for the pairwise posteriors (3000 posterior draws per pair). The division of the full dataset is shown in Table~\ref{pairwise}. 

\begin{table}[h]
\centering
\begin{tabular}{ccccccc}
\toprule
DDM parameter & Component & Distribution & Weight & Location/Shape & Scale & df \\
\midrule
$v$ & Dominant & Truncated normal & 0.85 & 1.76 & 1.51 & \\
    & Non-dominant & Lognormal & 0.15 & 1.53 & 0.54 & \\
$a$ & Dominant & Gamma & 0.76 & 11.69 & 0.12 & \\
    & Non Dominant & Gamma & 0.24 & 2.21 & 1.22 & \\
$z$ & & Truncated t & -- & 0.5 & 0.05 & 1.85 \\
$T_{er}$ & & Truncated t & -- & 0.44 & 0.08 & 1.32 \\
\bottomrule
\end{tabular}
\caption{DDM parameter distributions for sampling test data. The parameter $v$ and $a$ each follow a two-component mixture distribution, with the dominant and non-dominant components shown separately. The weights sum to 1 within each mixture. The other parameters have no mixture structure.}
\label{tab:simprior}
\end{table}

\begin{table}[ht!]
\renewcommand{\arraystretch}{1.4}
\centering
\begin{minipage}[t]{0.48\textwidth}
\centering
\begin{tabular}{ccccccccc}
\hline
Shard & Pair & $\hat{v}_1$ & $\hat{v}_2$ & $\hat{v}_3$ & $\hat{v}_4$ & $\hat{a}$ & $\hat{z}$ & $\hat{T}_{er}$ \\
\hline
$S_1$ & $(1,2)$  & $\hat{v}_1$ & $\hat{v}_2$ &       &       & $\hat{a}$ & $\hat{z}$ & $\hat{T}_{er}$ \\
$S_2$ & $(1,3)$  & $\hat{v}_1$ &       & $\hat{v}_3$ &       & $\hat{a}$ & $\hat{z}$ & $\hat{T}_{er}$ \\
$S_3$ & $(1,4)$  & $\hat{v}_1$ &       &       & $\hat{v}_4$ & $\hat{a}$ & $\hat{z}$ & $\hat{T}_{er}$ \\
$S_4$ & $(2,3)$ &     & $\hat{v}_2$ & $\hat{v}_3$ &       & $\hat{a}$ & $\hat{z}$ & $\hat{T}_{er}$ \\
$S_5$ & $(2,4)$ &     & $\hat{v}_2$ &       & $\hat{v}_4$ & $\hat{a}$ & $\hat{z}$ & $\hat{T}_{er}$ \\
$S_6$ & $(3,4)$ &  &  & $\hat{v}_3$ & $\hat{v}_4$ & $\hat{a}$ & $\hat{z}$ & $\hat{T}_{er}$ \\
\hline
 & & $\overline{v_1}$ & $\overline{v_2}$ & $\overline{v_3}$ & $\overline{v_4}$ & $\overline{a}$ & $\overline{z}$ & $\overline{T_{er}}$ \\
\hline
\end{tabular}
\label{pairwise1}
\end{minipage}
\hfill
\begin{minipage}[t]{0.48\textwidth}
\centering
\begin{tabular}{ccccccccc}
\hline
Shard &  Pair & $\hat{v}_1$ & $\hat{v}_2$ & $\hat{v}_3$ & $\hat{a}_1$ & $\hat{a}_2$ & $\hat{z}$ & $\hat{T}_{er}$ \\
\hline
$S_1$ & $(1,2)$ & $\hat{v}_1$ & $\hat{v}_2$ &       & $\hat{a}_1$ &       & $\hat{z}$ & $\hat{T}_{er}$ \\
$S_2$ & $(1,3)$ & $\hat{v}_1$ &       & $\hat{v}_3$ & $\hat{a}_1$ &       & $\hat{z}$ & $\hat{T}_{er}$ \\
$S_3$ & $(2,3)$ &   & $\hat{v}_2$ & $\hat{v}_3$ & $\hat{a}_1$ &       & $\hat{z}$ & $\hat{T}_{er}$ \\
$S_4$ & $(4,5)$ & $\hat{v}_1$ & $\hat{v}_2$ &       &       & $\hat{a}_2$ & $\hat{z}$ & $\hat{T}_{er}$ \\
$S_5$ & $(4,6)$ & $\hat{v}_1$ &       & $\hat{v}_3$ &       & $\hat{a}_2$ & $\hat{z}$ & $\hat{T}_{er}$ \\
$S_6$ & $(5,6)$ &       & $\hat{v}_2$ & $\hat{v}_3$ &       & $\hat{a}_2$ & $\hat{z}$ & $\hat{T}_{er}$ \\
\hline
 & & $\overline{v_1}$ & $\overline{v_2}$ & $\overline{v_3}$ & $\overline{a_1}$ & $\overline{a_2}$ & $\overline{z}$ & $\overline{T_{er}}$ \\
\hline
\end{tabular}
\label{pairwise2}
\end{minipage}

\caption{Pairwise shard division in settings 1 and 2 (left to right). 
Each row shows parameter estimates from the corresponding shard $S_q$ and which pair of conditions it contains. 
The same notation (e.g., $\hat v_k$, $\hat a$, $\hat z$) is reused across rows, but values are shard-specific. 
The bottom row reports the consensus estimate obtained from the relevant shards.}
\label{pairwise}
\end{table}

Posterior sampling with MCMC was performed in the R \texttt{Stan} \citep{stan}. We initially ran the analysis using four chains, each with 1000 warm-up (adaptation) iterations followed by 1000 sampling iterations, yielding 4,000 valid samples in total. For datasets that did not meet the convergence criteria (i.e., $\hat{R} < 1.01$ and effective sample size $N_{\text{eff}}$ > 400) \citep{vehtari2021improved}, we removed the fits for that dataset and ran the analysis again with same settings. All datasets for both full and the pairwise models satisfied the convergence criteria. 

\subsection*{Accuracy, uncertainty, and computational time assessments}

After posterior sampling, we obtained the consensus posterior via Equation~\ref{eq:consensus}, and subsequently the calibrated consensus posterior using importance sampling. We assessed the agreement between the full posterior and the calibrated consensus posterior. Mean absolute error (MAE) was used to measure the accuracy of posterior means for a single $\theta$ (a specific component of $\btheta$), where $\text{MAE}_{\theta} = \frac{1}{n} \sum|\hat{\theta}^r - \theta^r| $, and $\hat{\theta}^r$ is the posterior mean estimate for the $r$-th replicate and $\theta^r$ the true value. In addition, we report root mean squared error (RMSE), which accounts for residual variability by penalizing larger errors more heavily, defined as $\text{RMSE}_{\theta} = \sqrt{\frac{1}{n}\sum(\hat{\theta}^{r}-\theta^{r})^2}$. We also compared posterior uncertainty using the standard deviation (SD) of the calibrated consensus posterior and the full posterior.

Additionally, we report the computational time of estimation procedures for both ABI and MCMC sampling. For ABI, we reported the training time for the full and pairwise NPEs, as well as the average inference time across test datasets. For MCMC, we reported the average computation time per dataset and the total runtime for posterior sampling for both the full and pairwise models.

\subsection*{Setting 1 results }

\begin{figure}[!ht]
    \centering
    \includegraphics[width=1\linewidth]{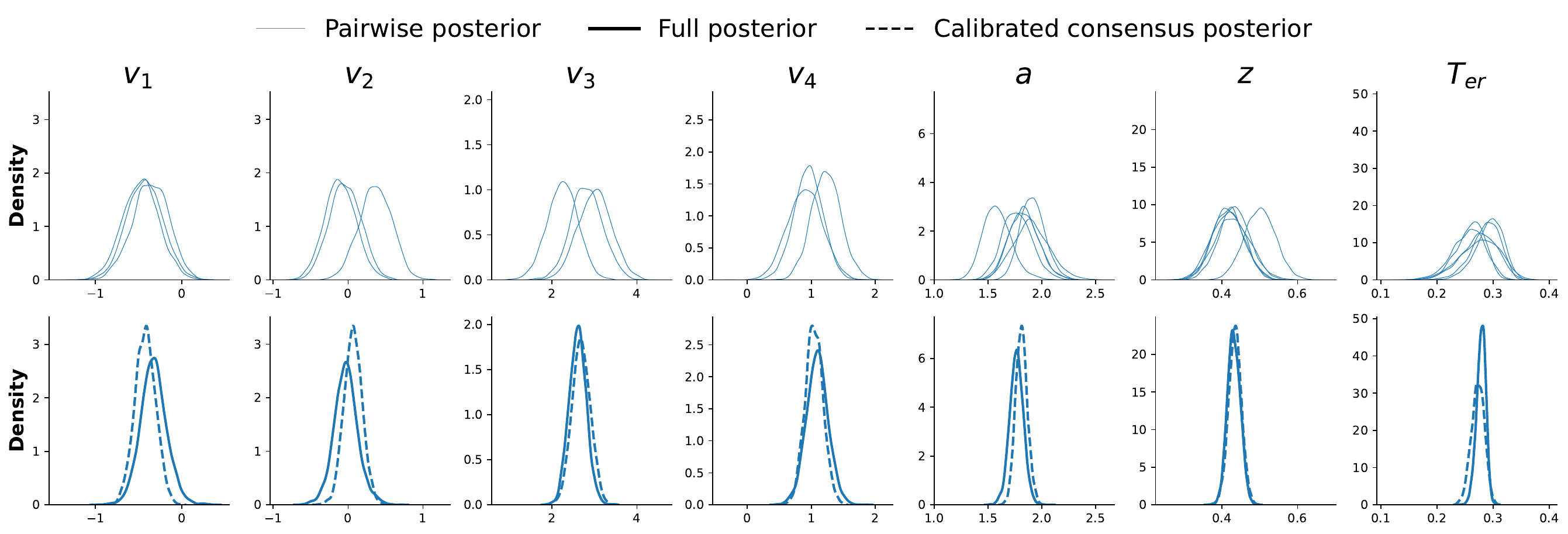}
    \caption{Example of posteriors of a single replicate when $n = 100$ in setting 1. The first row shows the marginal pairwise posterior for each parameter. The second row shows the full and calibrated consensus marginal posterior.}
    \label{fig:consensus_BF}
\end{figure}

We first selected one replicate to visualize the pairwise posterior, full posterior, and calibrated consensus posterior in Figure~\ref{fig:consensus_BF}. As expected, the marginal pairwise posteriors for the same parameter differ slightly in their means and SDs (first row of Figure~\ref{fig:consensus_BF}), as each pair contains different information. The calibrated consensus posterior aligns closely with the full posterior, suggesting that it can accurately approximate the full posterior for this replicate in Setting 1. Figure~\ref{fig:consensus_MCMC} shows the same pattern from the MCMC approach, with the calibrated consensus posterior closely matching the full posterior despite variability across pairwise posteriors. 

\begin{figure}
    \centering
    \includegraphics[width=1\linewidth]{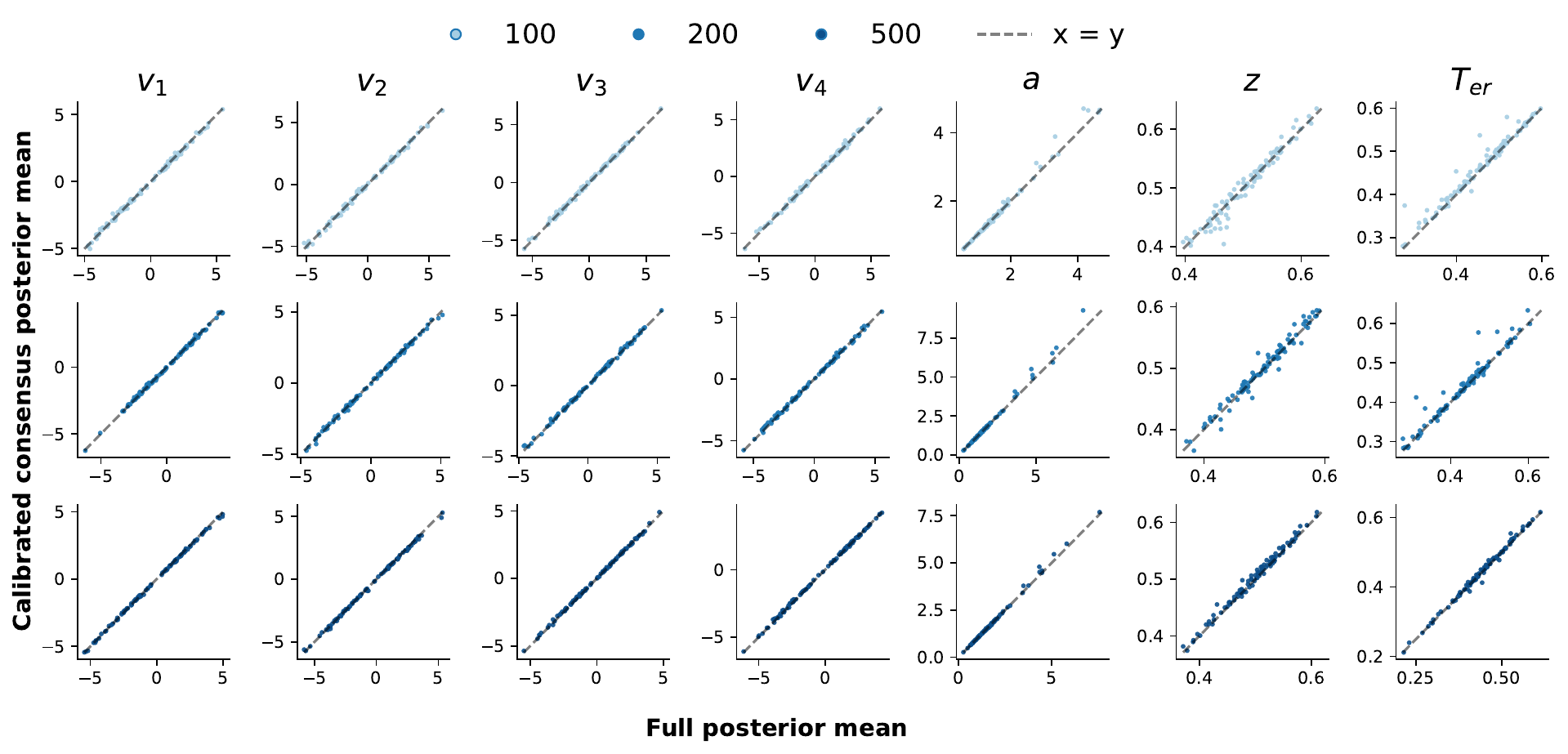}
    \caption{Full posterior mean vs. consensus posterior mean in setting 1. The x-axis indicates the full posterior mean for 100 replicates, while the y-axis shows their consensus posterior mean. The dashed line represents x = y.}
    \label{fig:posterior_mean_ABI_s1}
\end{figure}

As Figure~\ref{fig:consensus_BF} is only a single replicate, we systematically investigated whether the alignment between the calibrated consensus posteriors and the full model posteriors holds across replicates and values of $n$. Figure~\ref{fig:posterior_mean_ABI_s1} visualizes this by plotting the full posterior mean against the calibrated consensus posterior mean for all the replicates. Accuracy relative to the ground truth is reported in Table~\ref{tab:MAE_setting1}. Figure~\ref{fig:posterior_mean_ABI_s1} shows a strong agreement between the posterior means across all replicates, as they cluster closely around the diagonal line. The same pattern of alignment holds for MCMC-based estimation (Figure~\ref{fig:posterior_mean_MCMC_s1} in Appendix~\ref{app:mcmc_s1}), where the calibrated consensus posterior means closely align with the full posterior means around the diagonal line across all trial numbers.

\subsubsection*{Posterior mean accuracy}
We further present the MAE of the estimates. Table~\ref{tab:MAE_setting1} shows that, for the same $n$, the full model generally achieved lower MAE values (i.e., higher accuracy) than the pairwise estimation. However, the difference was minor and decreased as $n$ increased; in other words, larger datasets allow pairwise estimation to achieve the same level of accuracy as the full model. Interestingly, for $v_1$, $v_3$, and $T_{er}$ at $n = 500$, the full-model MAEs are slightly larger than the pairwise MAE. This may reflect the inherent difficulty of approximating the posterior of a complex stochastic model at high sample sizes, and we view it as an acceptable trade-off between amortization efficiency and estimation accuracy. 

Notably, Table~\ref{tab:MAE_setting1} reveals that MCMC sampling exhibited slightly higher accuracy than ABI in the full-model estimation, but yielded larger MAE values than its ABI counterpart in the pairwise estimation, particularly for drift rates when $n=100$ and $n=200$. This suggests that ABI provides a more accurate posterior mean in small-to-moderate sample settings. This pattern is likely attributable to the limited number of trials per shard: with four conditions each split into three parts, each pairwise fit received only approximately $n/3$ trials per condition, totaling around 66 trials at $n = 100$. This is likely to reflect an advantage of ABI: because the pairwise NPE was trained on datasets covering the full range of possible trial numbers, it has learned to approximate posteriors reliably even when only a small number of trials is available. Consistent with this interpretation, the MAE difference between MCMC and ABI narrows substantially at $n = 500$, particularly for the drift rate parameters, suggesting that the two methods perform more comparably when each pairwise fit receives sufficient trials. 

\begin{figure}[!ht]
\centering
\begin{subfigure}{1\linewidth}
    \centering
    \includegraphics[width=\linewidth]{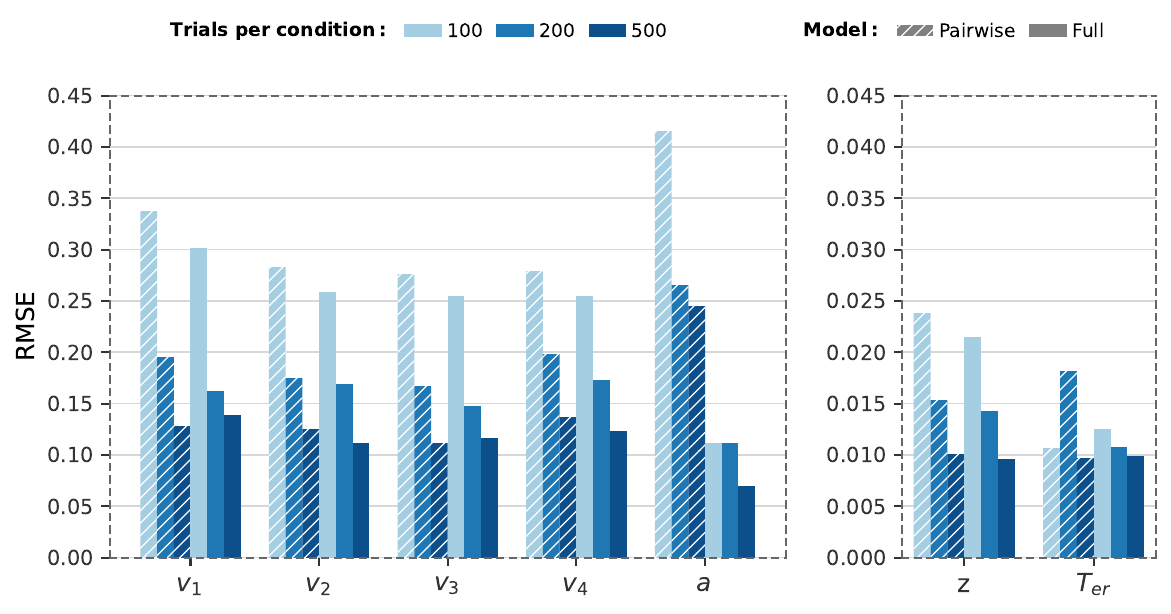}
    \caption{ABI estimation}
    \label{fig:rmse_s1_abi}
\end{subfigure}

\vspace{0.5cm}

\begin{subfigure}{1\linewidth}
    \centering
    \includegraphics[width=\linewidth]{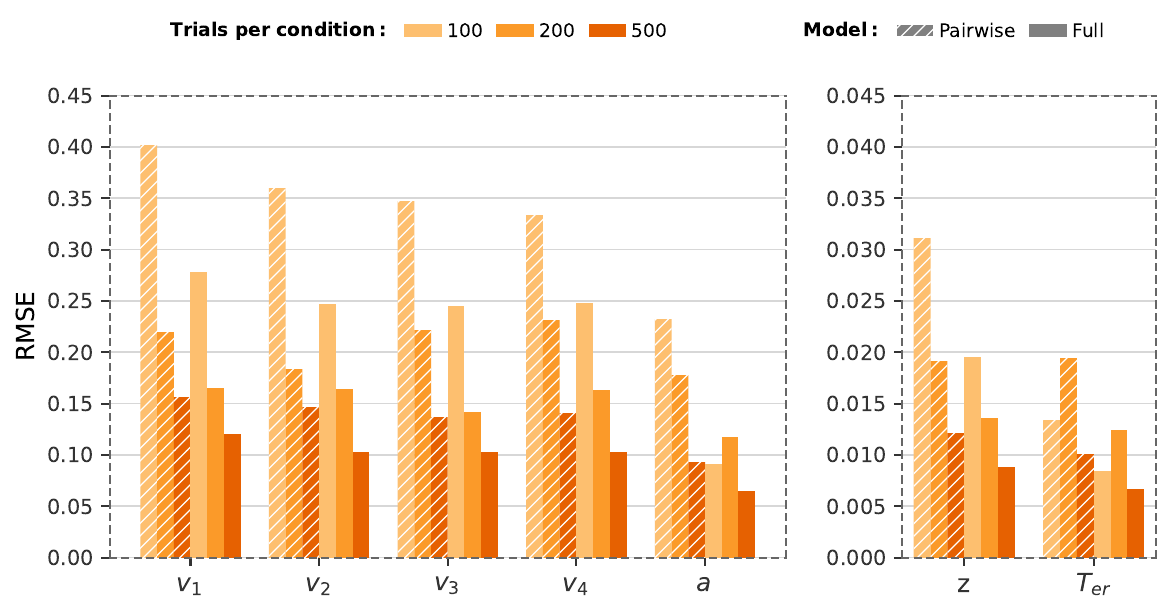}
    \caption{MCMC estimation}
    \label{fig:rmse_s1_mcmc}
\end{subfigure}

\caption{RMSE in setting 1. To enhance readability, the RMSE for response bias $z$ and non-decision time $T_{er}$ is in a smaller scale (see the y-axis ticks on the right.) }
\label{fig:rmse_setting1}
\end{figure}

The final metric used to evaluate accuracy is the RMSE per parameter for each estimation method (Figure~\ref{fig:rmse_setting1}). Unlike MAE, RMSE penalizes larger errors more heavily, making it more sensitive to outliers. The RMSE patterns and values obtained by the two approaches are highly comparable. For all drift rates ($v_1$ to $v_4$) and response bias ($z$), higher $n$ corresponds to lower RMSE, and pairwise estimation generally yields slightly higher RMSE than full-model estimation. The RMSE for boundary separation $a$ under pairwise ABI estimation is notably higher than under MCMC, suggesting that the neural network has greater difficulty approximating the posterior of $a$, possibly due to the limited number of trials available in each shard. Consistent with the MAE results, MCMC-based pairwise estimation exhibits larger RMSE than its ABI counterpart for drift rates ($v_1$ to $v_4$) at $n = 100$ and $n = 200$, and this difference decreases at $n = 500$. Overall, pairwise estimation achieves comparable accuracy to the full model for both ABI and MCMC.

\begin{table}[!ht]
\centering
\renewcommand{\arraystretch}{1.3}
\begin{tabular}{llcccccccc}
\hline
Method & Estimation & $n$ & $v_1$ & $v_2$ & $v_3$ & $v_4$ & $a$ & $z$ & $T_{er}$ \\
\hline 
\multirow{6}{*}{ABI} 
    &            & 100 & 0.253 & 0.222 & 0.206 & 0.210 & 0.106 & 0.018 & 0.007 \\
    &  Pairwise  & 200 & 0.145 & 0.137 & 0.131 & 0.160 & 0.075 & 0.012 & 0.008 \\
    &            & 500 & 0.092 & 0.095 & 0.083 & 0.101 & 0.061 & 0.008 & 0.004 \\
\cline{2-10}
    &            & 100 & 0.231 & 0.201 & 0.199 & 0.195 & 0.061 & 0.016 & 0.007 \\
    &  Full model& 200 & 0.121 & 0.131 & 0.113 & 0.141 & 0.056 & 0.011 & 0.006 \\
    &            & 500 & 0.102 & 0.090 & 0.089 & 0.088 & 0.035 & 0.008 & 0.005 \\
\hline

\multirow{6}{*}{MCMC}
    &            & 100 & 0.321 & 0.282 & 0.256 & 0.253 & 0.082 & 0.021 & 0.008   \\
    &  Pairwise  & 200 & 0.164 & 0.139 & 0.168 & 0.172 & 0.083 & 0.014 & 0.009  \\
    &            & 500 & 0.110 & 0.114 & 0.100 & 0.106 & 0.056 & 0.010 & 0.005  \\
\cline{2-10}
    &            & 100 & 0.215 & 0.185 & 0.189 & 0.185 & 0.060 & 0.015 & 0.005 \\
    & Full model & 200 & 0.121 & 0.130 & 0.107 & 0.132 & 0.072 & 0.010 & 0.006 \\
    &            &500  & 0.086 & 0.081 & 0.078 & 0.077 & 0.047 & 0.007 & 0.003\\
\hline

\end{tabular}
\caption{The MAE of pairwise and full model estimation with ABI in setting 1.}
\label{tab:MAE_setting1}
\end{table}

\subsubsection*{Posterior uncertainty}
\begin{figure}[!t]
    \centering
    \includegraphics[width=1\linewidth]{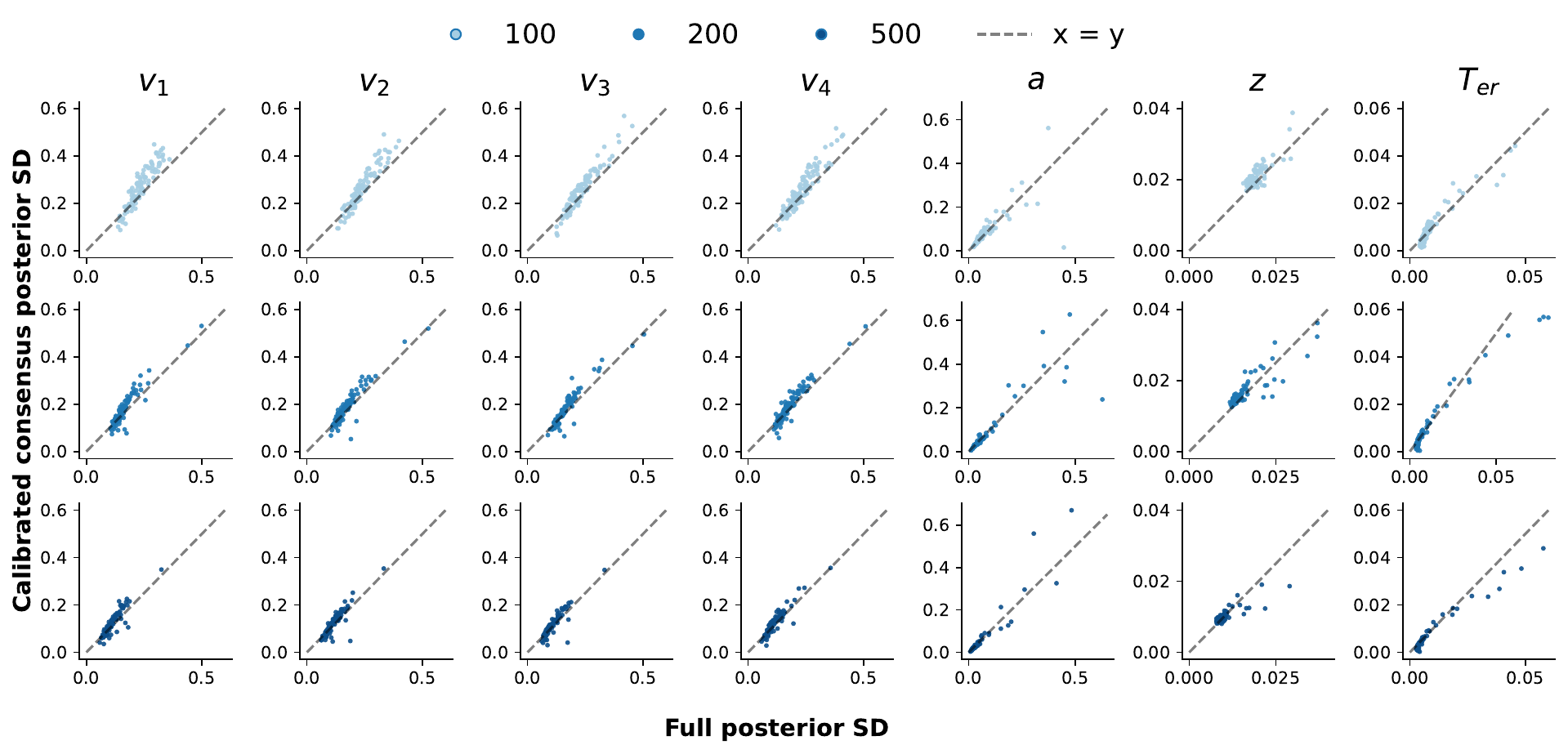}
    \caption{Full posterior SD vs. consensus posterior SD in setting 1. The x-axis indicates the full posterior SD for 100 replicates, while the y-axis shows their consensus posterior SD. The dashed line represents x = y.}
    \label{fig:posterior_SD_ABI_s1}
\end{figure}

We next investigated the posterior uncertainty of the estimates. Figure~\ref{fig:posterior_SD_ABI_s1} plots the full posterior SD against the calibrated consensus posterior SD for all replicates. Notably, the posterior SDs of the calibrated consensus posteriors are closely aligned with those of the full posterior, clustering tightly around the diagonal line ($x=y$) across all parameters and sample sizes. This indicates that the pairwise estimation framework not only recovers accurate point estimates but also faithfully captures posterior uncertainty. While the theoretical equivalence established in the early section guarantees that the pairwise decomposition targets the correct full posterior exactly, in practice the full pipeline introduces several sources of approximation error, including the neural network approximation, finite posterior draws, consensus MCMC, and importance sampling calibration. The close empirical agreement observed here therefore provides important practical validation that these approximation errors remain small and do not substantially distort the posterior uncertainty. The same pattern is observed for MCMC-based estimation, as shown in Figure~\ref{fig:posterior_SD_MCMC_s1} in the Appendix, further corroborating the robustness of the approach.

The average posterior SDs across all replicates are documented in Table~\ref{tab:sd_setting1}. Overall, both ABI and MCMC exhibited similar patterns: the calibrated consensus posterior SDs are slightly larger than the full model posterior SDs across all values of $n$, which is expected given that each pairwise fit uses only a subset of the data. The only exception is $T_{er}$ parameter under ABI, where pairwise estimation yielded slightly lower posterior SDs, suggesting mild overconfidence in uncertainty estimation for this parameter.

\begin{table}[!t]
\centering
\renewcommand{\arraystretch}{1.3}
\begin{tabular}{lcccccccccc}
\hline
Method & Estimation  & $n$ & $v_1$ & $v_2$ & $v_3$ & $v_4$ & $a$ & $z$ & $T_{er}$ \\
\hline
\multirow{6}{*}{ABI} 
    &             & 100 & 0.269 & 0.263 & 0.255 & 0.263 & 0.074 & 0.021 & 0.009 \\
    & Pairwise    & 200 & 0.177 & 0.185 & 0.179 & 0.192 & 0.071 & 0.016 & 0.009 \\
    &             & 500 & 0.122 & 0.123 & 0.113 & 0.123 & 0.049 & 0.010 & 0.006\\
\cline{2-10}
    &             & 100 & 0.235 & 0.238 & 0.234 & 0.240 & 0.073 & 0.019 & 0.009 \\
    & Full model  & 200 & 0.163 & 0.172 & 0.168 & 0.180 & 0.073 & 0.015 & 0.010 \\
    &             & 500 & 0.113 & 0.116 & 0.108 & 0.115 & 0.048 & 0.010 & 0.007 \\
\hline
\multirow{6}{*}{MCMC}
    &             & 100 & 0.260 & 0.259 & 0.251 & 0.260 & 0.068 & 0.020 & 0.009 \\
    & Pairwise    & 200 & 0.167 & 0.176 & 0.170 & 0.182 & 0.063 & 0.015 & 0.008 \\
    &             & 500 & 0.117 & 0.117 & 0.109 & 0.118 & 0.042 & 0.009 & 0.005\\
\cline{2-10}
    &             &100 & 0.228 & 0.226 & 0.220 & 0.227 & 0.064 & 0.019 & 0.007 \\
    & Full model  & 200 & 0.150 & 0.155 & 0.153 & 0.162 & 0.060 & 0.014 & 0.007 \\
    &             & 500 & 0.103 & 0.103 & 0.097 & 0.103 & 0.036 & 0.008 & 0.004 \\
\hline

\end{tabular}
\caption{The posterior SD of pairwise and full model estimation with ABI and MCMC in setting 1.}
\label{tab:sd_setting1}
\end{table}

\subsubsection*{Computational time}
Training the full-model NPE took approximately 17 seconds per epoch, resulting in a total of 28.33 minutes. Training the pairwise NPE took around 12 seconds per epoch, totaling 20 minutes. Obtaining 3,000 posterior draws from the full-model NPE required approximately 195 milliseconds for 100 simulated datasets, regardless of the trial number $n$. For the pairwise NPE, obtaining posterior samples from six shards across the 100 simulated datasets took approximately 730 milliseconds, also independent of $n$. Importantly, both inference times are independent of $n$, meaning that ABI scales well to larger datasets without any additional computational cost.

This stands in stark contrast to the MCMC approach, where computation time increased monotonically with the number of trials for both the full and pairwise models (see Table~\ref{tab:computation_time_MCMC_s1} in Appendix~\ref{app:mcmc_s1}). For the full model, the mean runtime per dataset increased from 7.50 seconds with 100 trials, to 15.61 seconds with 200 trials, to 39.28 seconds with 500 trials. This corresponds to total computation times of 12.50, 26.02, and 65.47 minutes, respectively. Although the mean runtime per fit was considerably shorter for the pairwise model than for the full model, the total computation time of the pairwise approach exceeded that of the full model. Unlike the ABI approach, MCMC requires a separate fitting procedure for each individual pair, resulting in a larger number of sequential model fits. In the pairwise model, the mean computation time increased from 1.41 seconds for 100 trials to 7.17 seconds for 500 trials, while the corresponding total computation times increased from 14.10 to 71.68 minutes. 

These results highlight an important asymmetry: the divide-and-conquer approach integrates naturally with ABI, adding only a small and fixed computational overhead regardless of $n$. For MCMC, however, the overhead of fitting each shard separately outweighs any gain from working with smaller shards, resulting in total computation times that exceed those of the full model. The computational benefits of the divide-and-conquer are therefore specific to the ABI framework.

\subsection*{Setting 2 results}

\begin{figure}[!ht]
    \centering
    \includegraphics[width=1\linewidth]{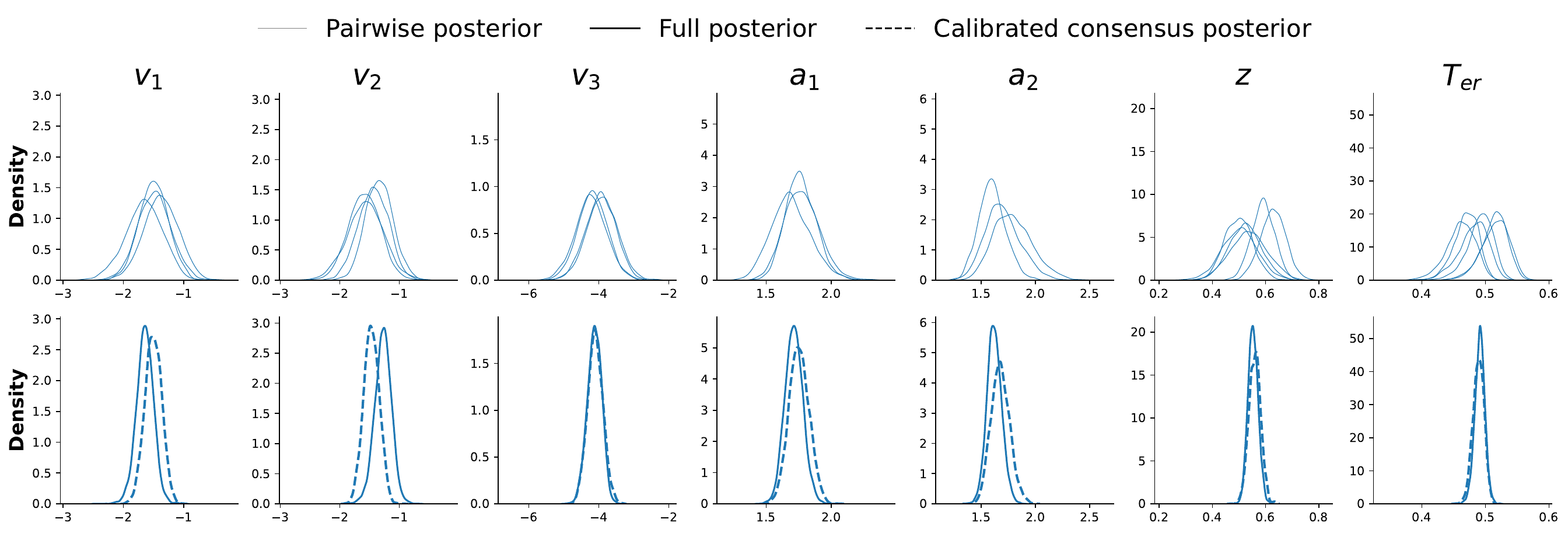}
    \caption{Example of posteriors for a single replicate when $n = 100$ in setting 2. The first row shows the marginal pairwise posterior for each parameter. The second row shows the full and calibrated consensus marginal posterior.}
    \label{fig:consensus_s2}
\end{figure}

As in setting 1, one replicate was randomly selected to illustrate the pairwise posteriors, full posterior, and calibrated consensus posterior in Figure~\ref{fig:consensus_s2}. The calibrated consensus posterior aligns closely with the full posterior across all parameters. The corresponding MCMC results (see Figure~\ref{fig:consensus_s2_MCMC} in Appendix~\ref{app:mcmc_s2}) show a similar overall pattern. 

\begin{figure}
    \centering
    \includegraphics[width=1\linewidth]{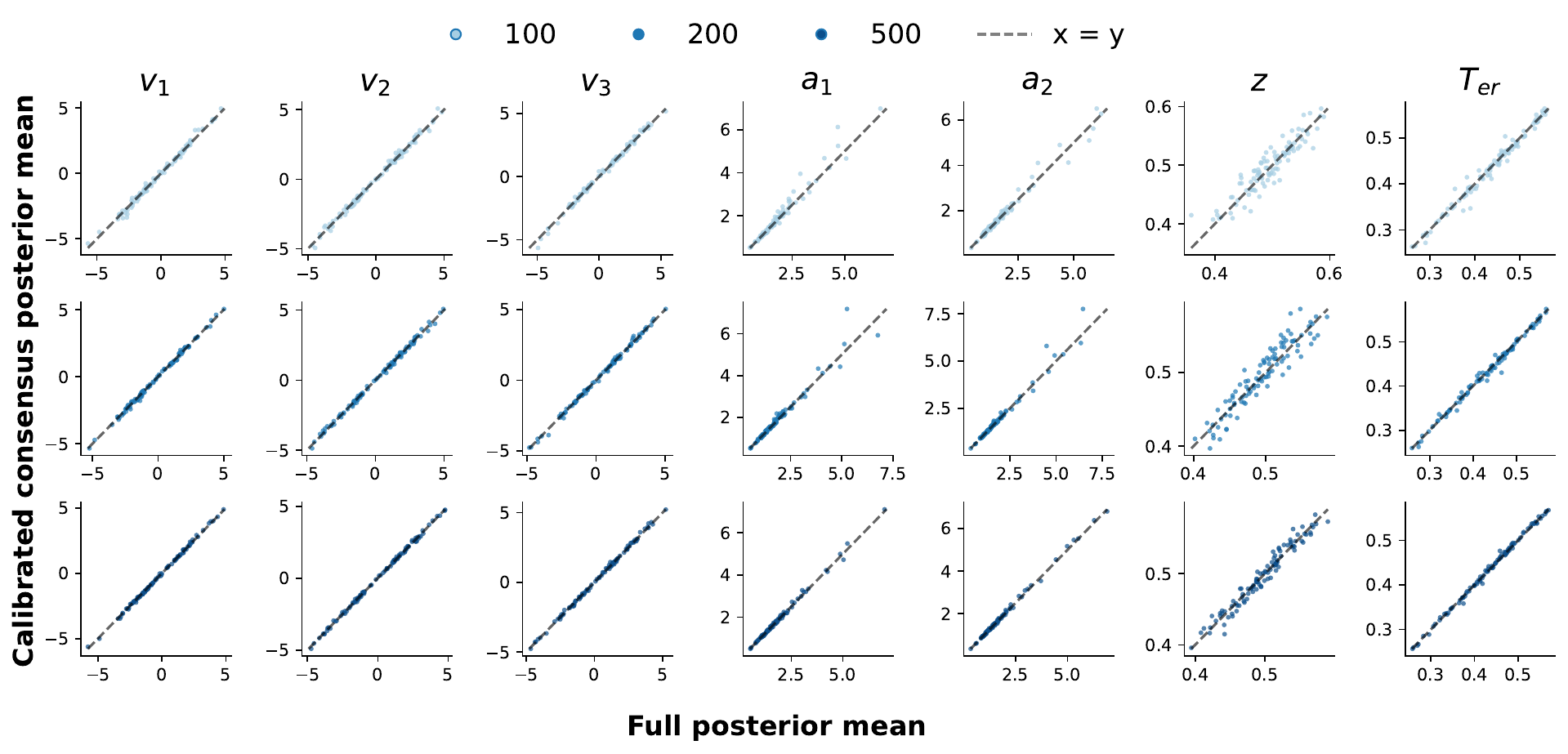}
    \caption{Full posterior mean vs. calibrated consensus posterior mean in setting 2. The x-axis indicates the full posterior mean, while the y-axis shows their consensus posterior mean. The dashed line represents x = y.}
    \label{fig:posterior_mean_ABI_s2}
\end{figure}

The calibrated consensus posterior means also align closely with the full posterior means across all parameters and sample sizes, with agreement improving as $n$ increases (Figure~\ref{fig:posterior_mean_ABI_s2}). Drift rates ($v_1$--$v_3$) 
and non-decision time ($T_{er}$) show particularly tight alignment, while boundary separation parameters ($a_1$, $a_2$) and starting point ($z$) exhibit slightly more spread at smaller sample sizes. The same pattern holds for MCMC-based estimation 
(Figure~\ref{fig:posterior_mean_MCMC_s2}).

\begin{table}[ht]
\centering
\renewcommand{\arraystretch}{1.3}
\begin{tabular}{lccccccccc}
\hline
Method & Estimation & $n$ & $v_1$ & $v_2$ & $v_3$ & $a_1$ & $a_2$ & $z$ & $T_{er}$ \\
\hline
\multirow{6}{*}{ABI} 
    &            & 100 & 0.196 & 0.200 & 0.182 & 0.127 & 0.127 & 0.021 & 0.007 \\
    &  Pairwise  & 200 & 0.119 & 0.121 & 0.117 & 0.105 & 0.080 & 0.014 & 0.004 \\
    &            & 500 & 0.079 & 0.073 & 0.091 & 0.045 & 0.048 & 0.008 & 0.003 \\
\cline{2-10}
    &            & 100 & 0.127 & 0.134 & 0.145 & 0.077 & 0.072 & 0.016 & 0.006 \\
    & Full model & 200 & 0.092 & 0.073 & 0.087 & 0.046 & 0.040 & 0.010 & 0.004 \\
    &            & 500 & 0.062 & 0.060 & 0.067 & 0.034 & 0.033 & 0.007 & 0.003 \\
\hline

\multirow{6}{*}{MCMC}
    &            & 100 & 0.166 & 0.146 & 0.165 & 0.110 & 0.118 & 0.017 & 0.006\\
    &   Pairwise & 200 & 0.109 & 0.097 & 0.102 & 0.082 & 0.070 & 0.012 & 0.003 \\
    &            & 500 & 0.070 & 0.063 & 0.068 & 0.067 & 0.066 & 0.006 & 0.003 \\
\cline{2-10}
    &            & 100 & 0.117 & 0.125 & 0.127 & 0.065 & 0.068 & 0.014 & 0.004\\
    & Full model & 200 & 0.079 & 0.062 & 0.071 & 0.050 & 0.046 & 0.009 & 0.003 \\
    &            & 500 & 0.054 & 0.048 & 0.055 & 0.042 & 0.045 & 0.005 & 0.002 \\
\hline
\end{tabular}
\caption{The MAE of pairwise and full model estimation with ABI and MCMC in setting 2.}
\label{tab:MAE_setting2}
\end{table}

\begin{figure}[!ht]
\centering

\begin{subfigure}{1\linewidth}
    \centering
    \includegraphics[width=\linewidth]{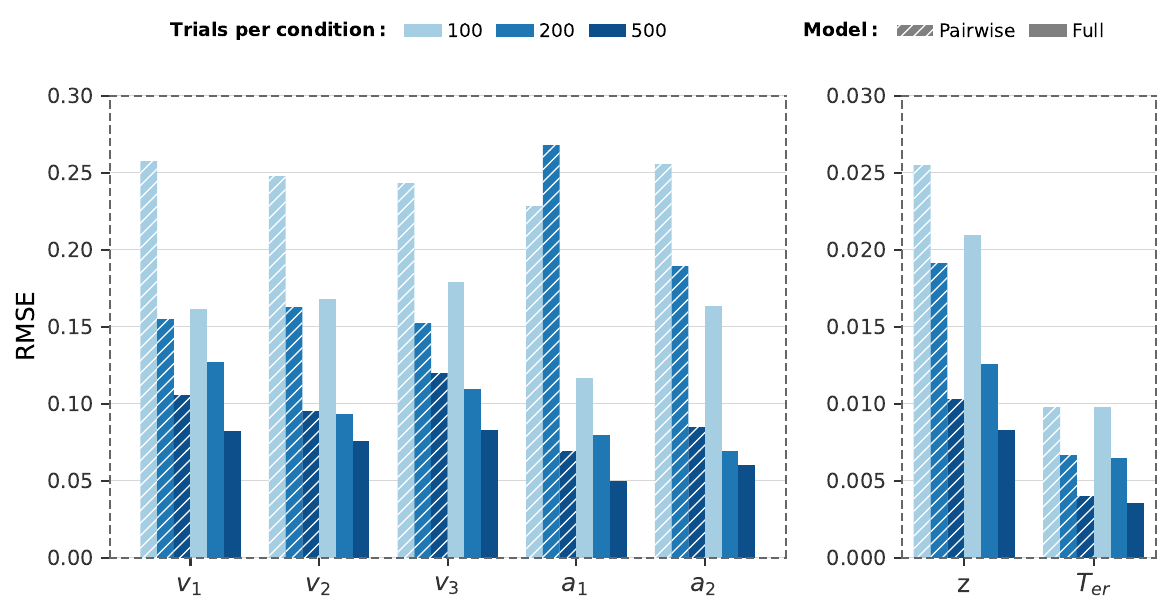}
    \caption{ABI estimation}
    \label{fig:rmse_s2_abi}
\end{subfigure}

\begin{subfigure}{1\linewidth}
    \centering
    \includegraphics[width=\linewidth]{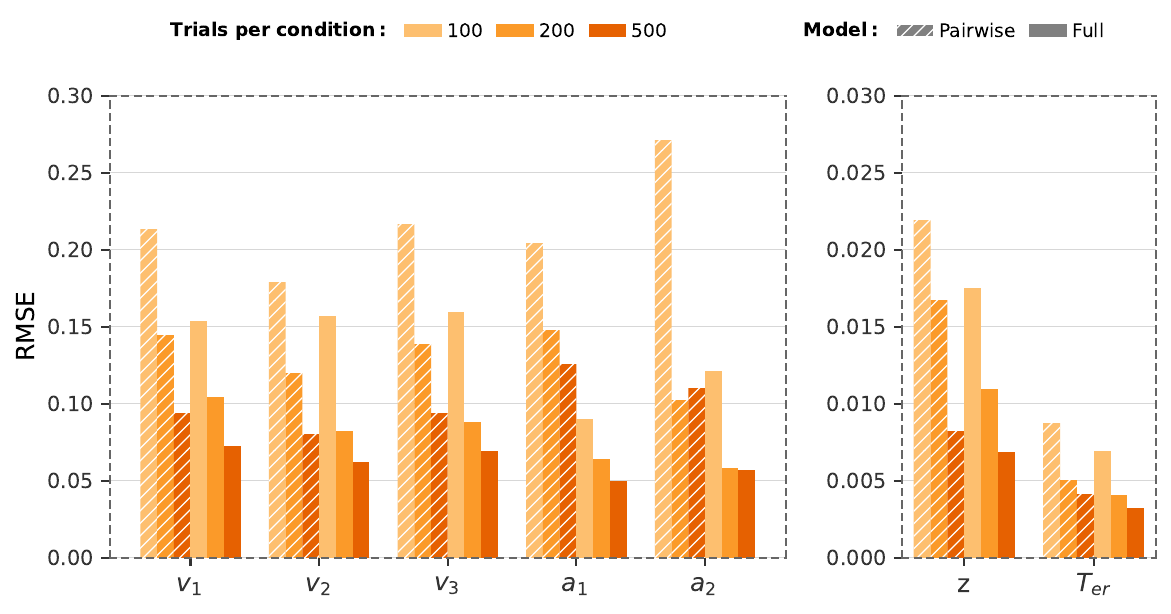}
    \caption{MCMC estimation}
    \label{fig:rmse_s2_mcmc}
\end{subfigure}

\caption{RMSE in setting 2. To enhance readability, the RMSE for response bias $z$ and non-decision time $T_{er}$ is in a smaller scale (see the y-axis ticks on the right.)}
\label{fig:rmse_setting2}
\end{figure}

\begin{figure}[ht!]
    \centering
    \includegraphics[width=1\linewidth]{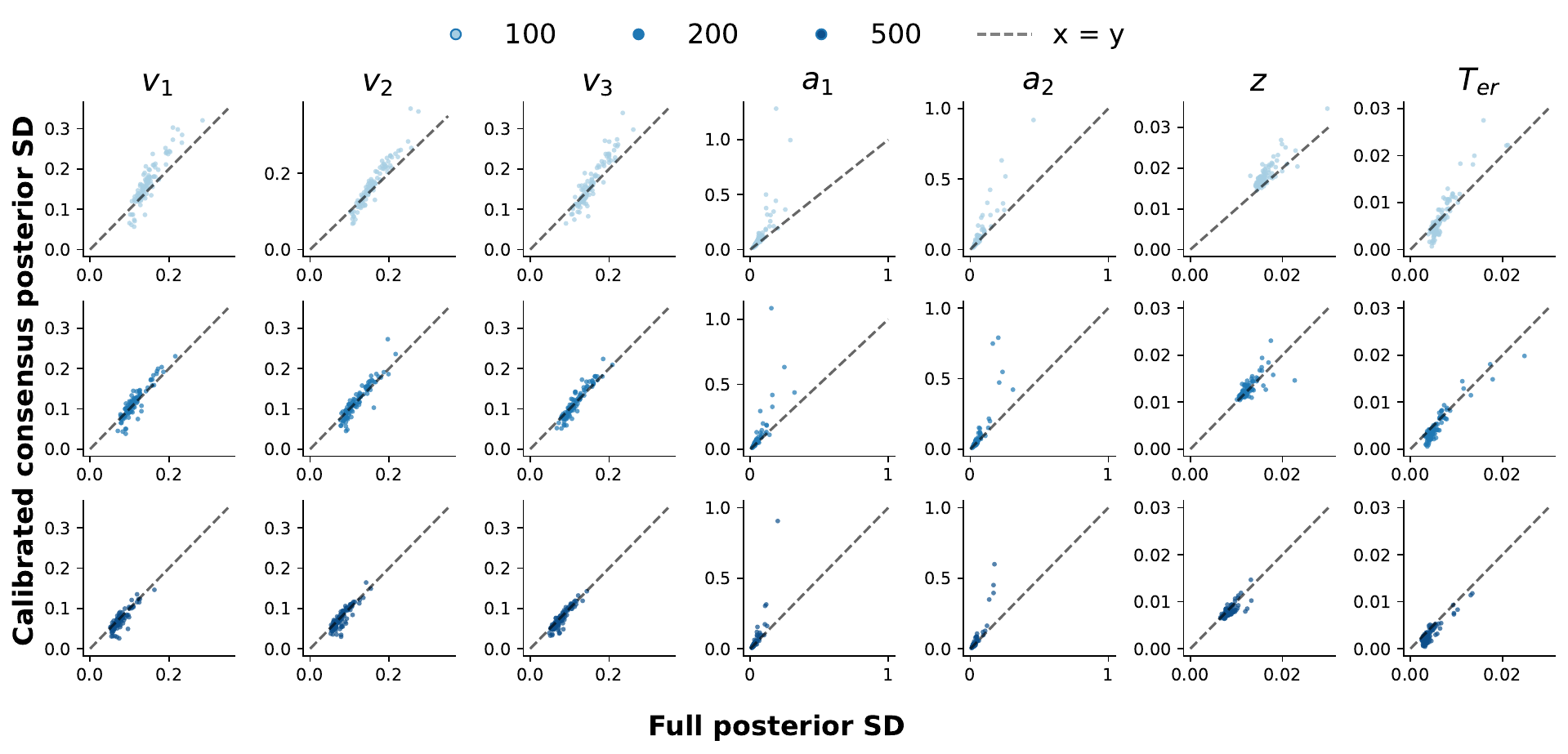}
    \caption{Full posterior SD vs. calibrated consensus posterior SD in setting 2. The x-axis indicates the full posterior SD for 100 replicates, while the y-axis shows their consensus posterior SD. The dashed line represents x = y.}
    \label{fig:posterior_SD_ABI_s2}
\end{figure}

\begin{table}[ht]
\centering
\renewcommand{\arraystretch}{1.3}
\begin{tabular}{lccccccccc}
\hline
Method & Estimation & $n$ & $v_1$ & $v_2$ & $v_3$ & $a_1$ & $a_2$ & $z$ & $T_{er}$ \\
\hline
\multirow{6}{*}{ABI}
    & Pairwise   &  100 & 0.164 & 0.168 & 0.163 & 0.120 & 0.103 & 0.018 & 0.007 \\
    &            &  200 & 0.113 & 0.118 & 0.115 & 0.084 & 0.078 & 0.013 & 0.005 \\
    &            &  500 & 0.070 & 0.075 & 0.073 & 0.052 & 0.050 & 0.008 & 0.003 \\
\cline{2-10}
    & Full model &  100 & 0.147 & 0.158 & 0.155 & 0.069 & 0.069 & 0.016 & 0.007\\
    &            &  200 & 0.111 & 0.118 & 0.115 & 0.053 & 0.052 & 0.013 & 0.006 \\
    &            &  500 & 0.076 & 0.078 & 0.078 & 0.035 & 0.035 & 0.008 & 0.004 \\
\hline
\multirow{6}{*}{MCMC}
    & Pairwise   &  100 & 0.155 & 0.162 & 0.157 & 0.095 & 0.108 & 0.017 & 0.006 \\
    &            &  200 & 0.108 & 0.114 & 0.111 & 0.072 & 0.068 & 0.012 & 0.004 \\
    &            &  500 & 0.068 & 0.072 & 0.071 & 0.046 & 0.044 & 0.008 & 0.003 \\
\cline{2-10}
    & Full model &  100 & 0.140 & 0.146 & 0.142 & 0.064 & 0.063 & 0.016 & 0.006\\
    &            &  200 & 0.098 & 0.103 & 0.100  & 0.044 & 0.043 & 0.011 & 0.004 \\
    &            &  500 & 0.062 & 0.065 & 0.064 & 0.028 & 0.027 & 0.007 & 0.003 \\
\hline
\end{tabular}
\caption{The posterior SD of pairwise and full model estimation with ABI and MCMC in setting 2.}
\label{tab:sd_setting2}
\end{table}

\subsubsection*{Posterior mean accuracy}
As in Setting 1, MAE decreases with increasing $n$ for both ABI and MCMC, and full-model estimation yields slightly lower MAE than pairwise estimation at the same $n$ 
(Table~\ref{tab:MAE_setting2}). Notably, the overprecision of pairwise ABI estimation observed at $n = 500$ in Setting 1 does not appear here, suggesting it may be specific to a particular design rather than a general limitation of the approach. Overall, MAE values between pairwise and full-model estimations remained highly comparable, confirming that the divide-and-conquer approach approximates full-model estimation well across both experimental designs. 

Consistent with the MAE results, RMSE values generally decreased with increasing $n$ for both ABI and MCMC (Figure~\ref{fig:rmse_setting2}). An exception is observed for $a_1$ under pairwise ABI and $a_2$ under pairwise MCMC, where RMSE does not decrease monotonically with $n$, consistent with the difficulty of recovering boundary separation parameters in pairwise estimation discussed above. Overall, full-model 
estimation yielded slightly lower RMSE than pairwise estimation across all parameters, though differences remained small, confirming that the divide-and-conquer approach recovers parameters with accuracy comparable to the full model.

\subsubsection*{Posterior uncertainty}
Same as in Setting 1, posterior SDs decrease with increasing $n$ for both methods, and drift rates ($v_1$--$v_3$) show broadly comparable SDs between pairwise and full model estimation (Table~\ref{tab:sd_setting2}, Figures~\ref{fig:posterior_SD_ABI_s2} 
and \ref{fig:posterior_SD_MCMC_s2}). For boundary separation parameters ($a_1$, $a_2$), pairwise estimates yields noticeably larger SDs than the full model for both ABI and MCMC, likely because each shard contains only a subset of the conditions that jointly inform $a$ in the full model. This over-dispersion is consistent across both inference methods, suggesting it reflects the information structure of the pairwise decomposition rather than a limitation of either method. As in Setting 1, mild overconfidence in $T_{er}$ at $n = 500$ is observed under pairwise ABI estimation.

\subsubsection*{Computational time}
Computational times for Setting 2 followed the same pattern as Setting 1. Training the full-model NPE took approximately 40 minutes, compared to 20 minutes for the pairwise NPE. ABI inference remained near-instantaneous and independent of $n$, requiring approximately 262 milliseconds for the full model and 635 milliseconds for the pairwise approach across 100 datasets. MCMC computation times again increased monotonically with $n$, with full-model runtimes ranging from 66.5 minutes at $n=100$ to 352.2 minutes at $n = 500$ (see Table~\ref{tab:computation_time_MCMC_s2}), further confirming that the computational advantages of pairwise ABI over MCMC are not specific to Setting 1.

\clearpage

\section{A real data example}
To demonstrate the practical utility of our divide-and-conquer approach, we applied it to the dataset from \cite{ratcliff_modeling_1998}, available as \verb+rr98+ in the \verb+RTDists+ R package \citep{rtdists2022}. This dataset provides a stringent test of the framework, as its high number of experimental conditions and imbalanced design go beyond the scenarios evaluated in the simulation studies. We fit data using NPE trained by our pairwise observation model and with MCMC sampling.

\begin{figure}[!ht]
\centering
\includegraphics[width=1\linewidth]{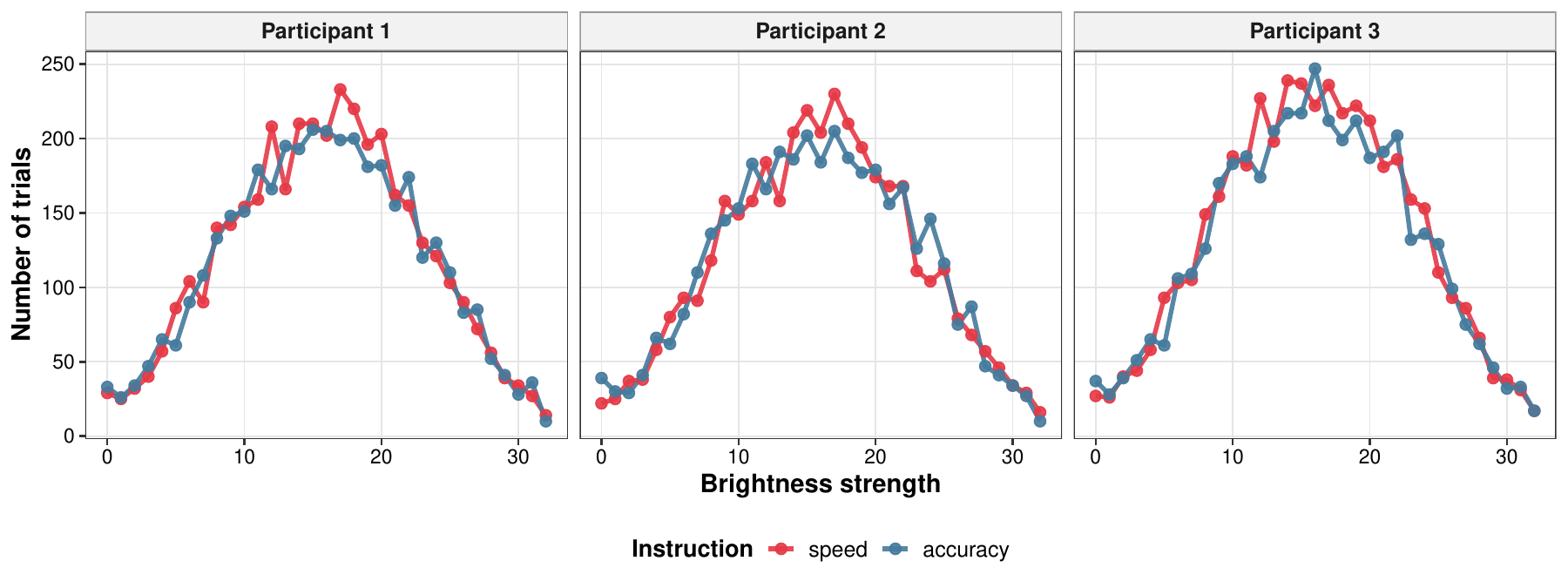}
\caption{Numbers of trials per condition. This figure shows the number of trials per brightness strength level under two instructions (speed or accuracy) for all participants.}\label{fig:trialnum}
\end{figure}

In this task, three participants were asked to decide whether the pixel arrays were ``bright'' or ``dark''. The experiment was a $33 \times 2$ within-subject design: brightness strength (0–100\% white pixels across 33 levels) and instruction (speed vs. accuracy). This setup targeted the speed–accuracy trade-off in decision-making. Each participant completed approximately 4,000 trials per instruction condition. The experimental design was imbalanced (see Figure~\ref{fig:trialnum}): conditions with more extreme brightness levels contain fewer trials. Data cleaning followed the original procedure of Ratcliff and Rouder (\citeyear{ratcliff_modeling_1998}), excluding trials with reaction times below 200 ms or above 2500 ms. After data cleaning, the minimum trial number in a condition was 10, while the maximum trial number within a condition was 247. We assumed different boundary separations for the speed and accuracy instructions, while the other parameters stayed constant across the two instruction levels.  

\begin{figure}[!ht]
\centering
\includegraphics[width=1\linewidth]{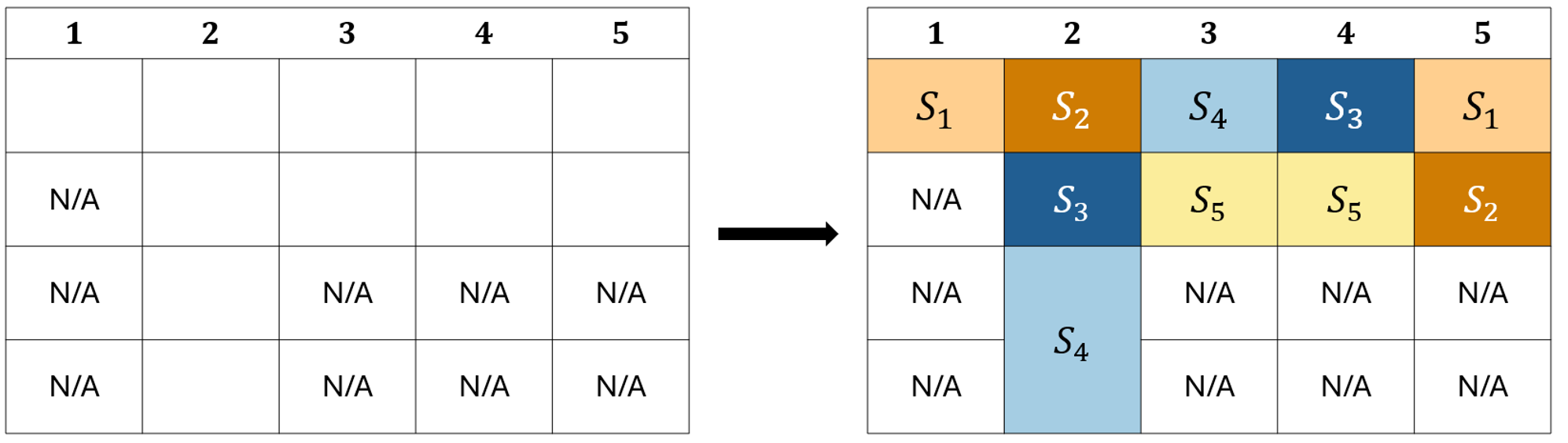}
\caption{The pairwise estimation with an imbalanced design. The left panel shows a data frame from a participant with 5 conditions. Each white square represents a mini data block, while ``N/A'' indicates no data available. The right panel illustrates the pairwise estimation: blocks sharing the same color are paired together, and the number in the upper-left corner of each square indicates the pairing order. Specifically, mini data blocks are paired from the most distant to the closest conditions.}\label{fig:imbalanced}
\end{figure}

One challenge here is that the trial numbers in some conditions are too low to be further partitioned. However, the pairwise estimation and consensus MCMC approach can still be applied to such a dataset by defining \textit{mini blocks} within a condition, where each block contains 10 trials, corresponding to the minimum trial number within a condition in \verb+rr98+ dataset. The only difference is that we cannot form pairs for every possible combination (see Figure~\ref{fig:imbalanced}), but the core approach remains valid.

Figure~\ref{fig:imbalanced} shows a hypothetical five-condition example, where each white square represents a mini block of data. As in the \verb+rr98+ dataset, conditions 1--5 correspond to brightness levels from 0 to 100\%. The contrast increases with the distance between conditions, so pairs from the most distant conditions are more informative and processed first. For adjacent pairs $(c, c+1)$, all remaining trials from condition $c$ are always consumed. Condition $c+1$ contributes all remaining trials only if no mini blocks remain in later conditions (i.e., condition $c+2$ onward); otherwise, it contributes a single mini block, preserving remaining trials for future pairings. This strategy ensures that all three core principles are satisfied: distant pairs are prioritized, every trial contributes to the analysis, and no trial is used more than once. After posterior sampling, we obtained the consensus posterior via Equation~\ref{eq:consensus}, and subsequently the calibrated consensus posterior using importance sampling.

The number of observations per condition in the training dataset varied from 10 to 250 trials to accommodate the wide range of trial numbers of \verb+rr98+ dataset. Training was performed on an NVIDIA Tesla H100-SXM2-32GB GPU, with each epoch taking 7 seconds and a total training time of 11.67 minutes. After the training was finished, we input pairs of observations into the NPE and obtained 1000 posterior samples for each pair. This was completed on CPU (two 32-core AMD EPYC 9334 processors) and took an average of 85 ms for each pair. The fitting involved aggregating information across 1056 pairs of estimation, and for each participant the fitting takes around 90s. 

For each pairwise fitting, we ran four parallel MCMC chains with 1000 warm-up iterations followed by 1000 sampling iterations, resulting in 4000 valid posterior samples per pair. For fits that did not meet convergence criteria ($\hat{R}$ > 1.01 or $N_{\text{eff}}$ < 400), we ran the model again until there were no warnings and convergence criteria were achieved. The average runtime per pairwise fit was approximately 0.47 seconds. The total numbers of pairwise fits were 369, 357, and 407 for the three subjects, respectively, resulting in total sampling times of 2.69, 2.68, and 3.46 minutes.

\begin{figure}[!ht]
\centering
\includegraphics[width=1\linewidth]{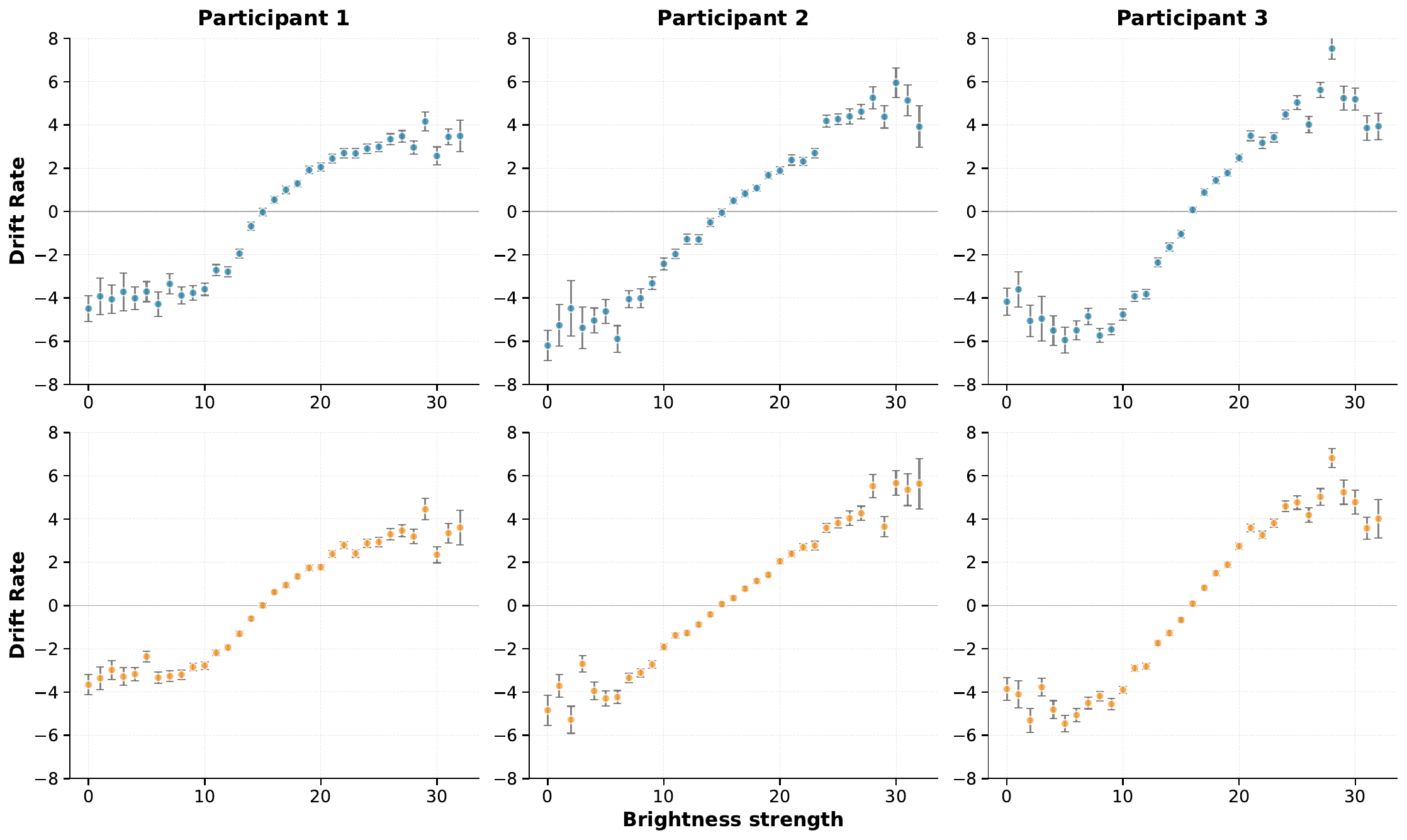}
\caption{Estimated drift rates of 33 brightness strength levels. The blue scatters represent the calibrated consensus posterior means obtained from the ABI approach, whereas the orange scatter points represent the posterior means obtained from the MCMC approach. The error bars show $\pm$ calibrated consensus posterior standard deviation, quantifying the uncertainty. }
\label{fig:driftestimates}
\end{figure}

The estimates of drift rates is shown in Figure~\ref{fig:driftestimates}. The trends in estimates were aligned with the experimental design: higher the brightness strength was associated with larger the posterior mean drift rates. As conditions with more extreme drift rates had fewer observations, their drift rate estimates were associated with higher posterior standard deviations, therefore greater uncertainty in the estimates. Importantly, the overall patterns of drift rate estimates were highly consistent between the ABI and MCMC approaches. Although minor differences between the two approaches were observable for conditions located toward the lower and higher ends of the brightness-strength range, these differences likely reflect the smaller number of observations available in those conditions, which resulted in greater estimation uncertainty. 

For reference, Figure~\ref{fig:Full_MCMC_real.data.example} in the Appendix presents the posterior drift rates estimated from the full MCMC model across brightness strength levels for all three participants. This suggests that the consensus approach successfully captures the core structure of the evidence accumulation process. However, there are differences between the pairwise and full model estimations. The full MCMC posterior yields considerably tighter credible intervals compared to the consensus posteriors, consistent with the larger uncertainty pattern shown in simulation studies, where consensus posterior SDs tend to exceed those of the full model. Second, the consensus posteriors tend to produce more extreme drift rate estimates at the tails of the brightness-strength range compared to the full model, showing more negative values at low brightness levels and more positive values at high brightness levels for each participant. 

The estimates of other parameters from the two approaches can be seen in Table~\ref{tab:estimates}. As expected, the boundary separation under speed instruction is lower than that under accuracy instruction, as participants were more cautious when instructed to respond as accurately as possible. For all participants, the decision-making process started from the middle of the two boundaries. The estimates of $T_{er}$ were highly similar across participants, possibly due to the hard cutoff of 200 ms applied during data cleaning. 

Table~\ref{tab:realdata_full_mcmc} presents the corresponding parameter estimates from the full MCMC model for reference. Although the full MCMC estimates of drift rates show the same pattern as the weighted consensus posteriors, the full MCMC tends to estimate larger boundary separation compared to both consensus posterior estimation approaches. For instance,the full MCMC estimates $a_{\text{accuracy}}$ (participant 1) as 2.092, while the ABI and MCMC's calibrated consensus estimates were 1.653 and 1.620 respectively. The SDs of the full MCMC estimates are also notably smaller than those of the consensus posteriors, consistent with the over-dispersion pattern observed in the simulation studies. 

This discrepancy between pairwise and full model estimation is larger than observed in the simulation studies, likely reflecting the fundamental information insufficiency inherent in the mini block approach when trial numbers per condition are small. With as few as 10 trials per mini-block, the shard posteriors may deviate substantially from the Gaussian form assumed by consensus MCMC, potentially introducing additional bias in the combination step. It is worth noting, however, that the \verb+rr98+ dataset represents an extreme case that serves as a particularly challenging test of the framework. In experimental designs where each condition contains a more reasonable number of trials, the pairwise posteriors would be approximately Gaussians, the calibrated consensus posterior would be more accurate, and the discrepancy between pairwise and full model estimation would be expected to diminish substantially, consistent with the pattern observed across sample sizes in the simulation studies.

\begin{table}[h]
\centering
\begin{tabular}{lccccc}
\toprule
Method & Participants & $a_\text{speed}$ & $a_\text{accuracy}$ & z & $T_{er}$ \\
\midrule
\multirow{3}{*}{ABI}
& 1 & 0.739 (0.012) & 1.653 (0.028) & 0.444 (0.005) & 0.221 (0.001)\\
& 2 & 0.728 (0.013) & 1.888 (0.033) & 0.496 (0.006) & 0.213 (0.001)\\
& 3 & 0.860 (0.015) & 1.546 (0.028) & 0.453 (0.005) & 0.238 (0.002) \\
\midrule
\multirow{3}{*}{MCMC}
& 1 & 0.763 (0.007) & 1.620 (0.020) & 0.467 (0.004) & 0.225 (0.001)\\
& 2 & 0.745 (0.008) & 1.873 (0.022) & 0.500 (0.004) & 0.214 (0.000)\\
& 3 & 0.877 (0.010) & 1.539 (0.019) & 0.518 (0.004) & 0.246 (0.000)\\
\bottomrule
\end{tabular}
\caption{Estimated DDM parameters (excluding drift rates) for all participants. Values represent the mean of calibrated consensus posteriors with standard deviations in parentheses, derived from ABI and MCMC methods.}
\label{tab:estimates}
\end{table}

Overall, this example illustrates both the strength and the limitations of the pairwise approach: it scales successfully to complex real-world designs that would be intractable for full-model ABI, while performing most reliably when each condition contributes a sufficient number of trials.

\section{Discussion}

This paper was motivated by a fundamental limitation of ABI: while it offers near-instantaneous parameter estimation, the trained networks are inherently tied to the experimental design they were trained on. A network trained on data from a three-condition experiment cannot be directly applied to a four- or five-condition experiment, as the dimensionality and structure of the input and output differ from what the network has learned. In practice, this means that every new experimental design requires a new network, which undermines the efficiency gains that make ABI attractive in the first place.

A first contribution of this paper is the divide-and-conquer framework addressing the aforementioned limitation. By decomposing any experimental design into condition pairs, a single pairwise NPE can be applied across diverse designs without retraining. This is the core contribution: transforming a design-specific tool into a generalizable inference engine. Consensus MCMC followed with an importance sampling calibration step provides the practical machinery for combining pairwise posteriors into a coherent approximation of the full posterior.

A second contribution of this work concerns the scalability of ABI to high-dimensional inference problems. Full-model ABI is fundamentally limited by the expressiveness of the neural network: as the number of conditions grows, the posterior becomes increasingly high-dimensional, placing severe demands on both the summary network's capacity to compress the data and the inference network's ability to approximate the posterior geometry. During the development of this framework, we consistently encountered biased posterior estimates when training full-model NPEs, attributable to the high dimensionality of the inference problem. Using conditional flow matching with a time-conditioned MLP substantially improved performance, rendering the bias undetectable in simulation-based calibration checks. This experience ultimately motivated the pairwise approach: rather than seeking more expressive architectures, we reduced the dimensionality of the inference problem itself. By keeping inference always in a low-dimensional pairwise space, our approach largely sidesteps this limitation, offering reliable posterior estimation regardless of the number of experimental conditions.

Nevertheless, the divide-and-conquer approach is not universally applicable. The approach assumes i.i.d. observations across trials and conditions, an assumption that is violated when, for example, the drift rate on a given trial depends on the outcome of the previous trial. In such cases, a specialized sequential summary network would be required, and the pairwise estimator would produce biased results. Researchers should therefore carefully consider whether the i.i.d. assumption is tenable before applying this framework.

Taken together, these findings establish pairwise estimation with ABI as a scalable and principled alternative to full-model inference, substantially expanding the feasibility of rapid Bayesian estimation in high-dimensional cognitive modeling while clearly delineating the conditions under which its assumptions must be carefully considered.

\bibliography{ref}

\appendix 

\begin{appendices}

\setcounter{figure}{0}
\setcounter{table}{0} 
\renewcommand{\thefigure}{A\arabic{figure}}
\renewcommand{\thetable}{A\arabic{table}}

\section{ABI workflow}
\label{app:abi} 
\begin{figure}[!ht]
\centering
\includegraphics[width=1\linewidth]{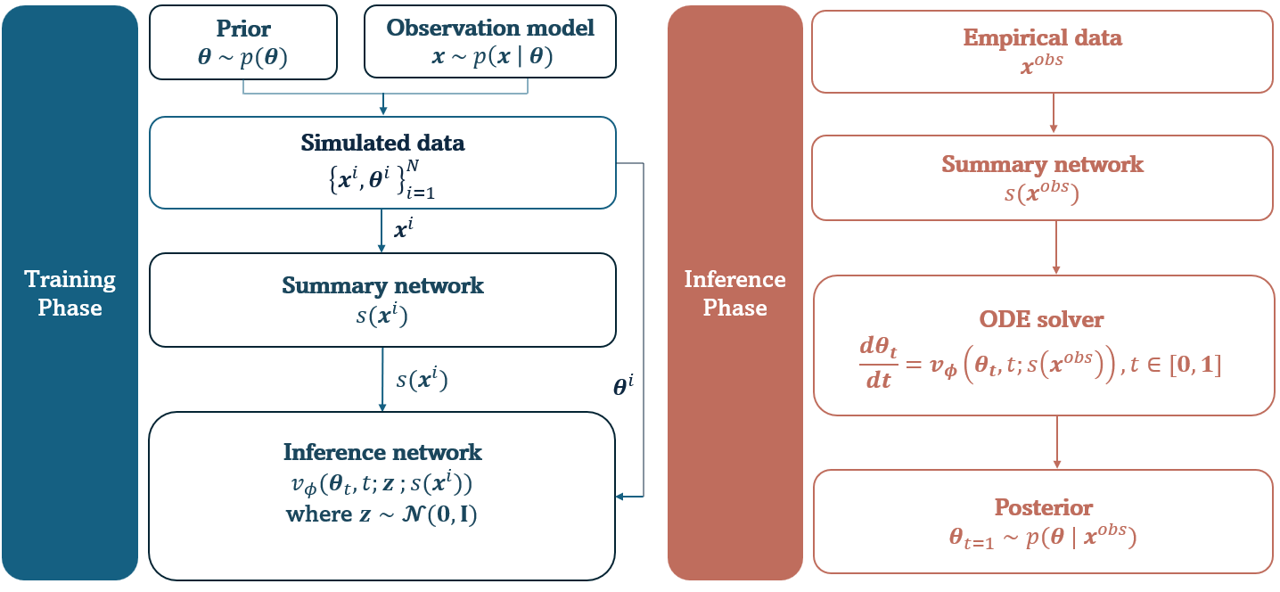}
\caption{\label{fig:BayesFlowworkflow} The basic workflow of conditional flow matching in ABI for posterior estimation. Parameters and data are simulated from a prior and an observation model. These simulations are used as training data for the summary and inference networks that jointly learn a vector field that transforms random noise to posterior samples from $t=0$ to $t=1$. Once trained, the networks can instantly sample from the posterior for any observed data by an ODE solver.}
\end{figure}

A basic workflow is depicted in Figure~\ref{fig:BayesFlowworkflow}. The first step is to generate synthetic data as training data. One first defines an observational model $\bx \sim p(\bx \mid \btheta)$ with a prior $\btheta \sim p(\btheta)$ and, where $\btheta \in \mathbb{R}^D$. Parameters are simulated from the prior and passed to the observational model to generate training data. The simulated parameters and data $\{\bx^i, \btheta^i\}_{i=1}^I$ are used to train two neural networks: a summary and an inference network. The summary network transforms each data set $\bx^i$ into fixed-size \textit{approximately sufficient} summary statistics $s(\bx^i)$, where $s(\cdot)$ represents the transformation applied by the summary network. The inference network approximates complicated posterior distributions via algorithms such as conditional flow matching \citep[CFM,][]{lipmanflow}, one of the most expressive architectures that is currently available. With CFM, the inference network learns a time-dependent vector field $v_t$ that can map a predefined simple base distribution $\boldsymbol{z}$ (e.g., a spherical Gaussian) to a complex target distribution, where $t \in [0, 1]$:
\begin{equation*}
    \btheta \sim p(\btheta \mid \bx) \Longleftrightarrow \btheta = \psi_1 (\boldsymbol{z};\,s(\bx)) \quad \text{with} \quad \boldsymbol{z} \sim \mathcal{N}(\mathbf{0}, \mathbf{I}), \label{eq:flow_matching}
\end{equation*}
where $\psi_1$ is the flow map generated by integrating the conditional vector field $v_t$ from $t=0$ to $t = 1$.

Upon convergence, the vector field $v_\phi$ is learned, drawing samples from the approximate posterior $q(\btheta \mid s(\bx^{\text{obs}}))$ involves solving an ordinary differential equations (ODE). Given an observation $\bx^{\text{obs}}$, we sample an initial state from the base distribution, $\btheta_{t=0} = \boldsymbol{z} \sim \mathcal{N}(\mathbf{0}, \mathbf{I})$, and simulate the continuous-time dynamics governed by the learned vector field:
\begin{equation*}
    \frac{\mathrm{d}\btheta_t}{\mathrm{d}t} = v_\phi(\btheta_t, t; s(\bx^{\text{obs}})), \quad t \in [0, 1].
\end{equation*}
The final state at $t=1$, denoted as $\btheta_{t=1}$, constitutes a sample from the targeted posterior distribution. In practice, this integration is performed using a numerical ODE solver (e.g., the Euler method).

\clearpage
\section{Importance sampling}
\label{app:Importance_samp}
We first obtain the marginal consensus posterior draws $\btheta_1,\dots, \btheta_L$:
$$
    \btheta_1,..., \btheta_L \sim p(\btheta \mid \bx) = \frac{\prod_{q=1}^{Q} p(S_q \mid \btheta)p(\btheta)^Q}{\int \prod_{q=1}^{Q} p( S_q \mid \btheta)p(\btheta)^Qd\btheta},
$$
and we denote the integral $C = \int \prod_{q=1}^{Q} p(S_q \mid \btheta)p(\btheta)^Qd\btheta$. The goal here is to obtain the weighted marginal consensus posterior draws $\btheta^\star_1,\dots,\btheta^\star_l$ that comes from a new, target distribution:
\begin{equation*}
    \btheta^\star_1,..., \btheta^\star_L \sim p^\star(\btheta \mid \bx) = \frac{\prod_{q=1}^{Q} p(S_q \mid \btheta)p(\theta)}{\int \prod_{q=1}^{Q} p(S_q \mid \btheta)p(\btheta)d\btheta},
\end{equation*}
and we denote the integral $C^* = \int \prod_{q=1}^{Q} p(S_q \mid \btheta)p(\btheta)d\btheta$. To this end, we employ importance sampling. The ratio between the target and consensus posteriors is given by:
\begin{equation*}
    \frac{p^\star(\btheta \mid \bx)}{p(\btheta \mid \bx)} = 
    \frac{C}{C^\star} \cdot
    \frac{\prod_{q=1}^{Q} p(S_q \mid \btheta)p(\btheta)}
    {\prod_{q=1}^{Q} p(S_q \mid \btheta)p(\btheta)^Q} = 
    \frac{C}{C^\star}p(\btheta)^{1-Q}.
\end{equation*}

Thus, up to a normalizing constant, the importance weights are
\begin{equation*}
    w(\btheta) \propto p(\btheta)^{1-Q}.
\end{equation*}

This induces a weighted empirical approximation of the target posterior:
\begin{equation*}
    p^\star(\btheta \mid \bx) \approx \sum_{l=1}^L\tilde{w}_l\delta_{\btheta_{l}}(\btheta),
\end{equation*}
where $\tilde{w}_l = \frac{w(\btheta_l)}{\sum_{l=1}^{L}w(\btheta_l)}$ are normalized importance weights and $\delta_{\btheta_l}$ denotes a point mass at $\btheta_l$.



\clearpage
\section{MCMC results from setting 1}
\label{app:mcmc_s1}

\begin{figure}[!ht]
    \centering
    \includegraphics[width=1\linewidth]{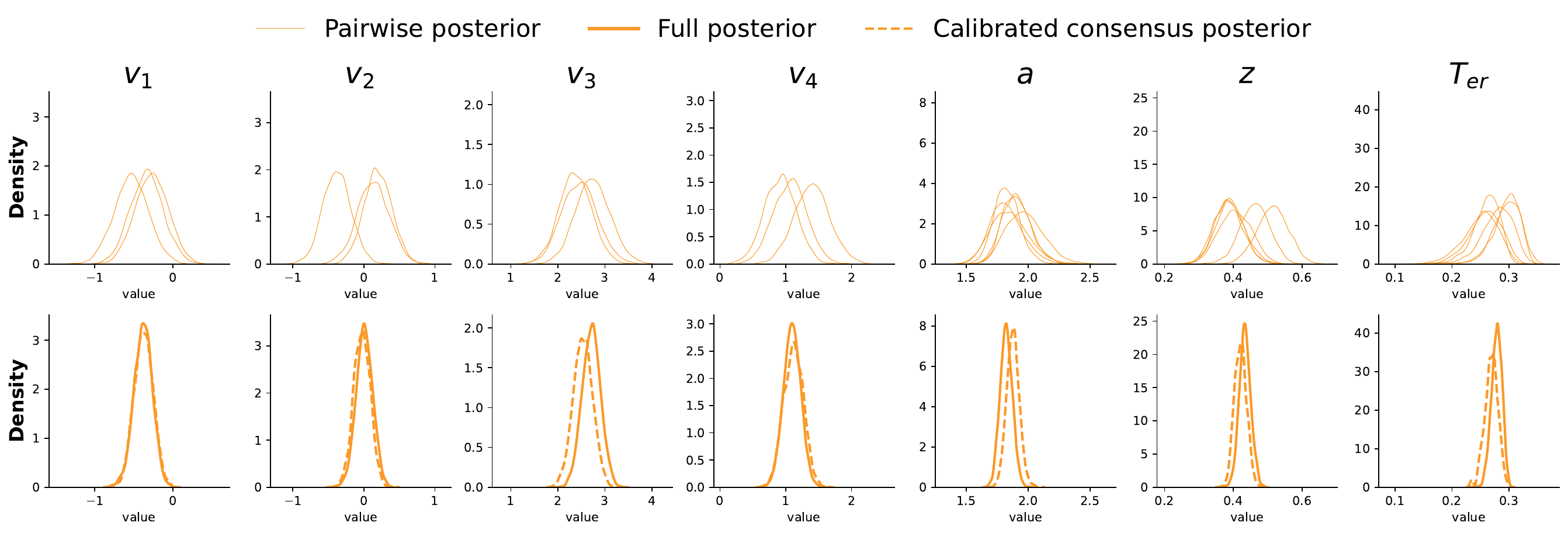}
    \caption{Example of posteriors of a replicate when $n = 100$ in setting 1. The first row shows the marginal pairwise posterior for each parameter. The second row shows the full and consensus marginal posterior.}
    \label{fig:consensus_MCMC}
\end{figure}

\begin{figure}[!ht]
    \centering
    \includegraphics[width=1\linewidth]{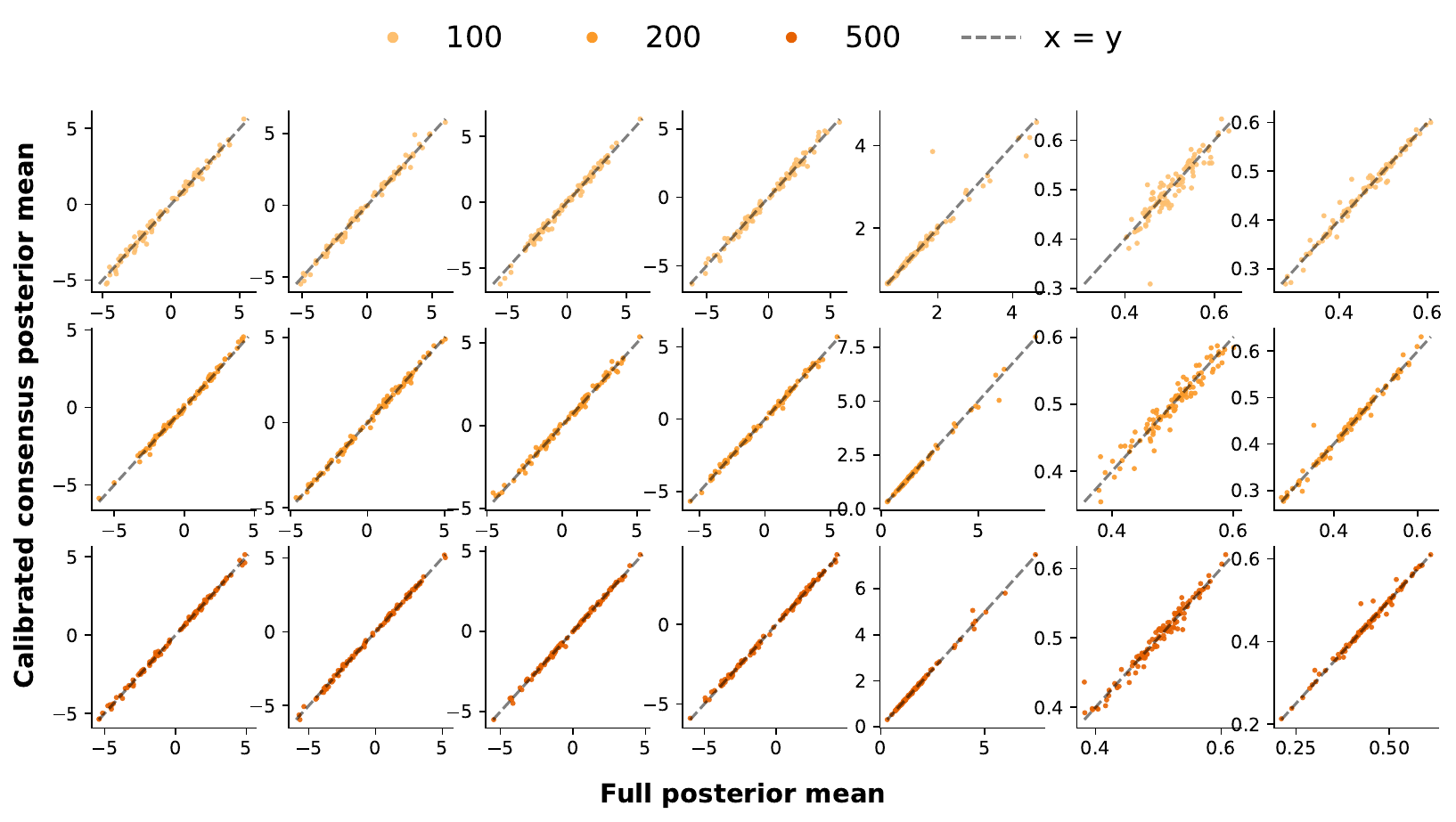}
    \caption{Full posterior mean vs. consensus posterior mean in setting 1. The x-axis indicates the full posterior mean for 100 replicates, while the y-axis shows their consensus posterior mean. The dashed line represents x = y.}
    \label{fig:posterior_mean_MCMC_s1}
\end{figure}

\begin{figure}[H]
    \centering
    \includegraphics[width=1\linewidth]{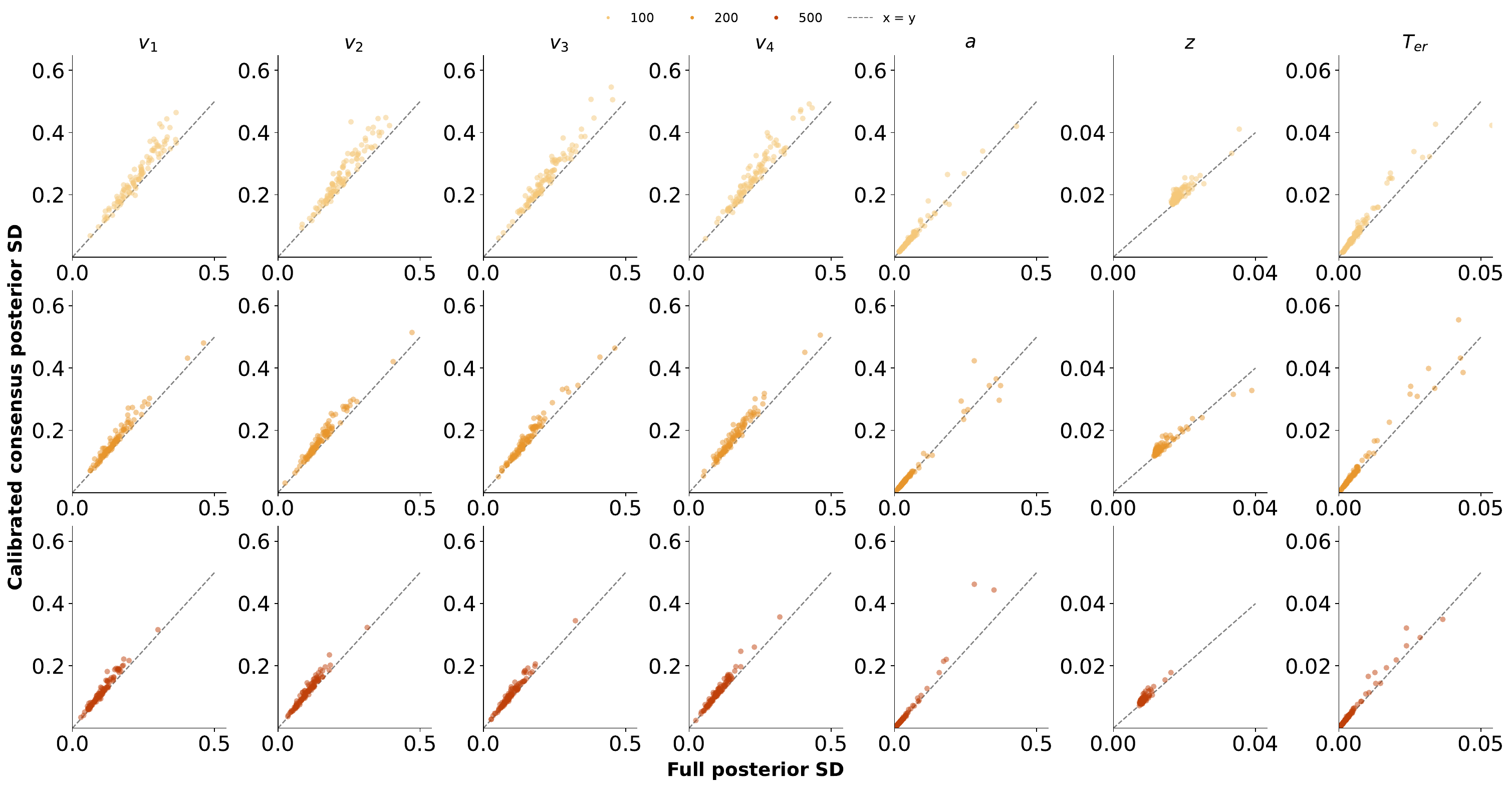}
    \caption{Full posterior SD vs. consensus posterior SD in setting 1. The x-axis indicates the full posterior SD for 100 replicates, while the y-axis shows their consensus posterior SD. The dashed line represents x = y.}
    \label{fig:posterior_SD_MCMC_s1}
\end{figure}

\begin{table}[H]
      \centering
      \begin{tabular}{lrcccc}
      \hline
       & \multicolumn{2}{c}{Full model} & \multicolumn{2}{c}{Pairwise model} \\
      Trials & Mean time (s) & Total time (min) & Mean time (s) & Total time (min) \\
      \hline
            100 & 7.50  & 12.50 & 1.41 & 14.10 \\
            200 & 15.61 & 26.02 & 2.75 & 27.48 \\
            500 & 39.28 & 65.47 & 7.17 & 71.68 \\
      \hline
\end{tabular}
\caption{Computation time for the full model and pairwise model as a function of the number of trials from MCMC approach for setting 1.}
\label{tab:computation_time_MCMC_s1}
\end{table}

\clearpage
\section{MCMC results from setting 2}
\label{app:mcmc_s2}
\begin{figure}[ht]
    \centering
    \includegraphics[width=1\linewidth]{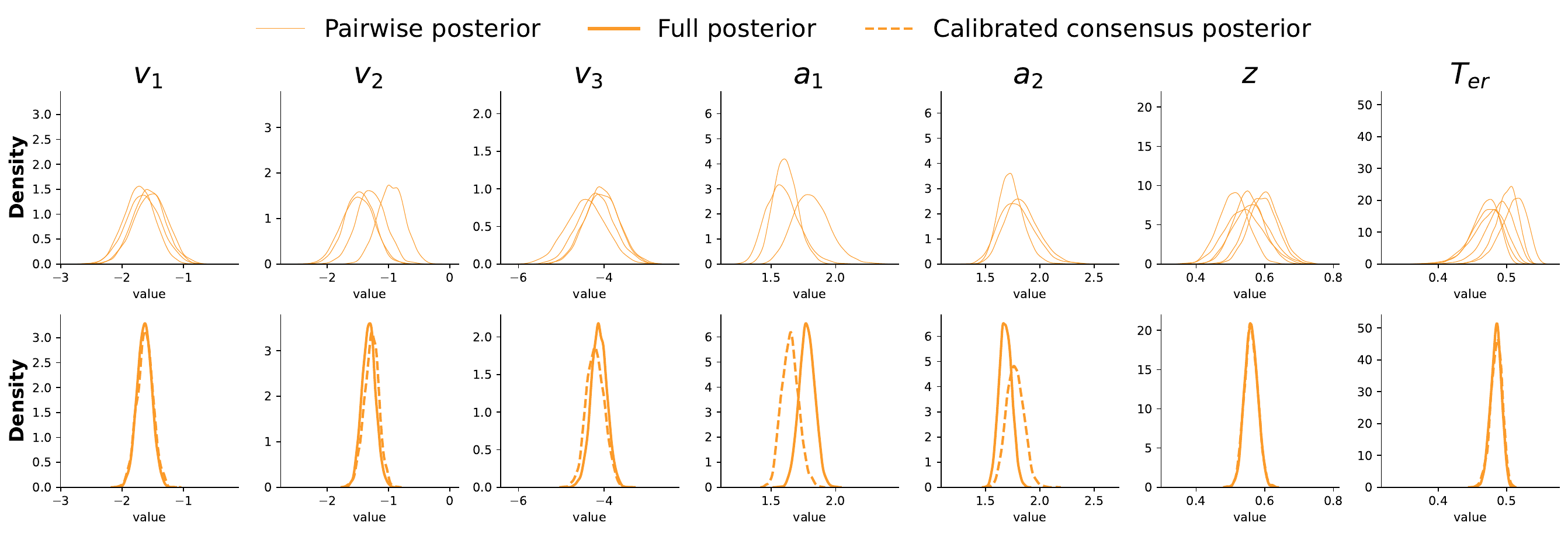}
    \caption{Example of posteriors of a replicate when $n = 100$ in setting 2. The first row shows the marginal pairwise posterior for each parameter. The second row shows the full and consensus marginal posterior.}
    \label{fig:consensus_s2_MCMC}
\end{figure}

\begin{figure}[!ht]
    \centering
    \includegraphics[width=1\linewidth]{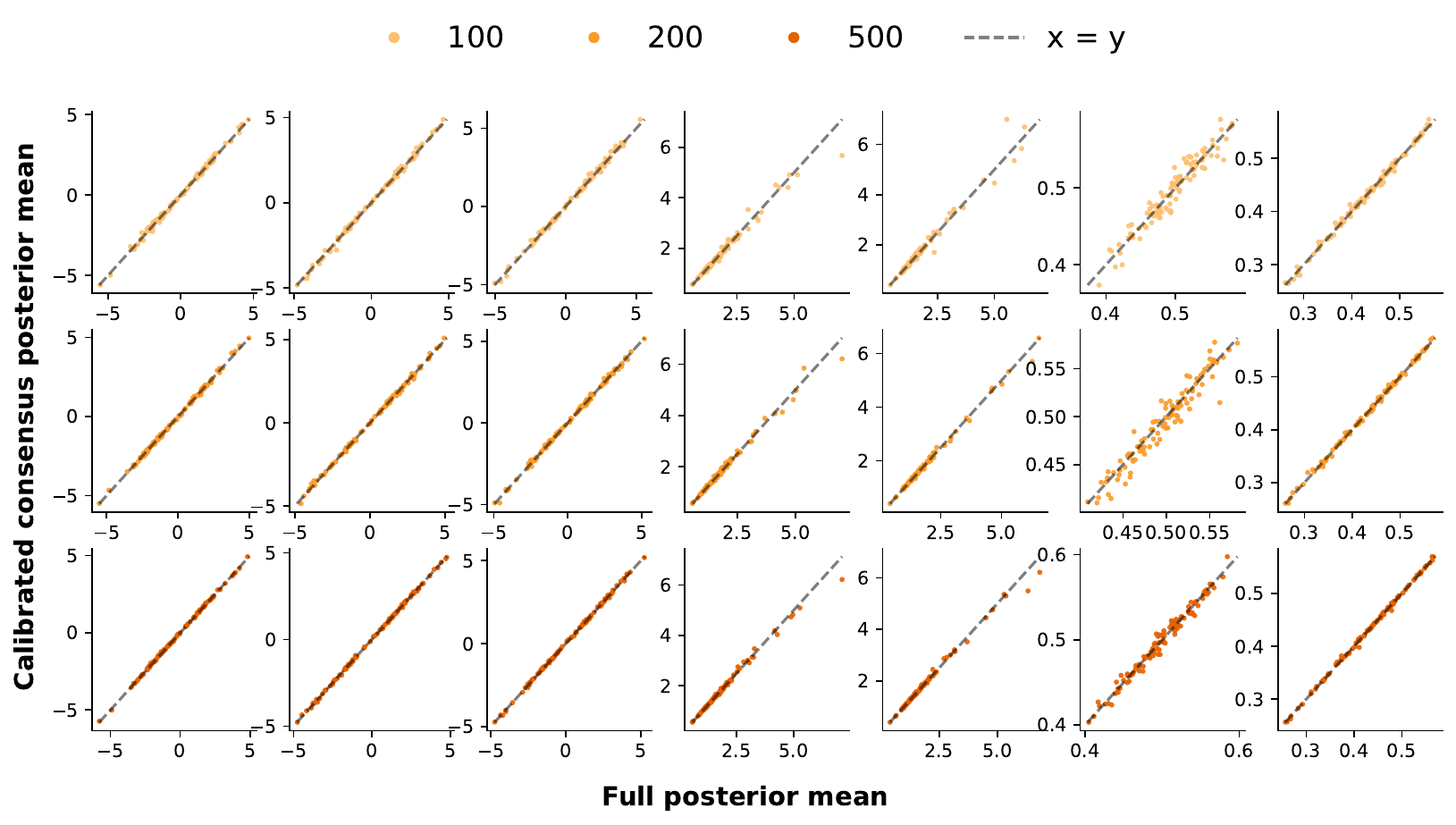}
    \caption{Full posterior mean vs. consensus posterior mean in setting 2. The x-axis indicates the full posterior mean, while the y-axis shows their consensus posterior mean. The dashed line represents x = y.}
    \label{fig:posterior_mean_MCMC_s2}
\end{figure}

\begin{figure}[!ht]
    \centering
    \includegraphics[width=1\linewidth]{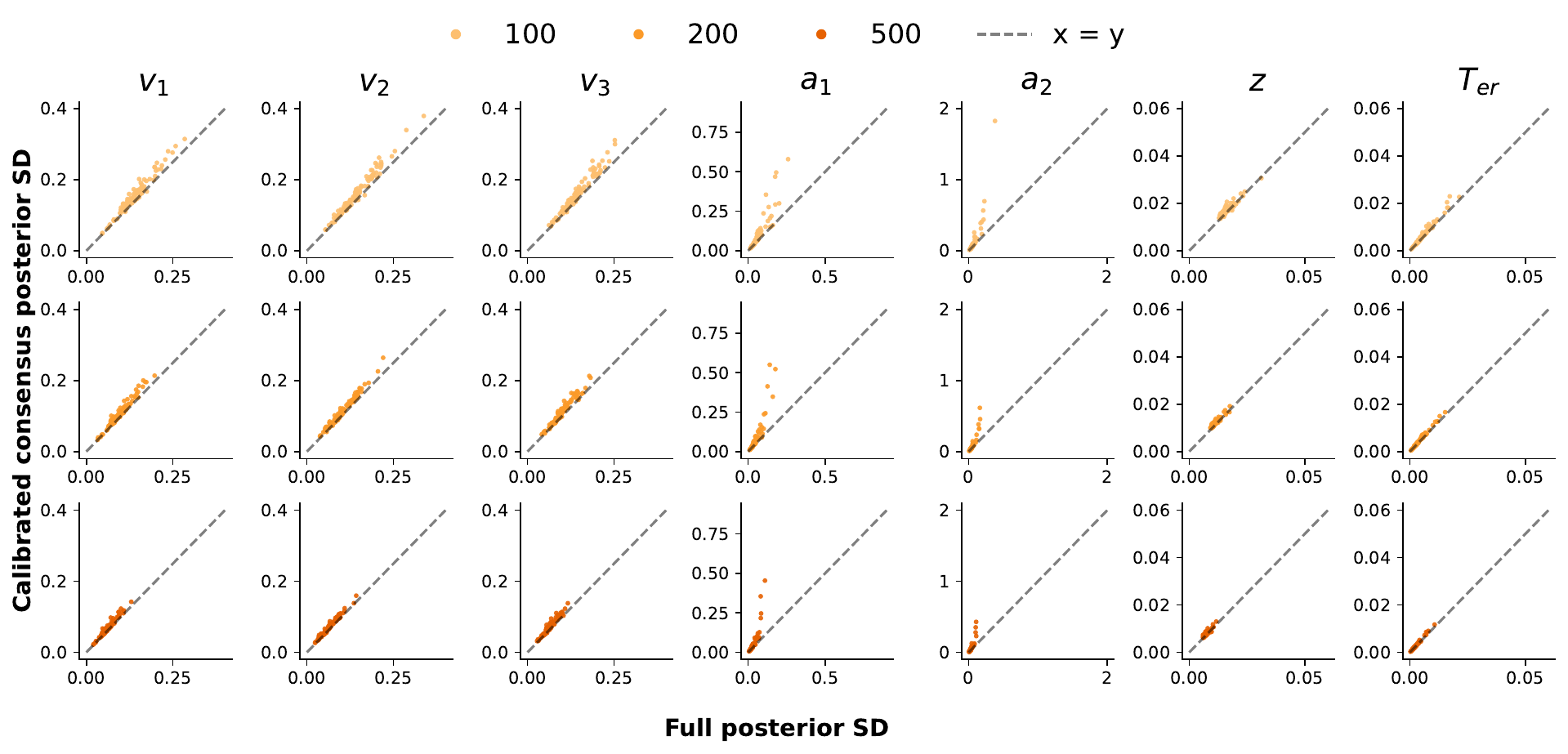}
    \caption{Full posterior SD vs. consensus posterior SD in setting 2. The x-axis indicates the full posterior SD, while the y-axis shows their consensus posterior SD. The dashed line represents x = y.}
    \label{fig:posterior_SD_MCMC_s2}
\end{figure}

\begin{table}[!t]
      \centering
      \begin{tabular}{lrcccc}
      \hline
       & \multicolumn{2}{c}{Full model} & \multicolumn{2}{c}{Pairwise model} \\
       Trials & Mean time (s) & Total time (min) & Mean time (s) & Total time (min) \\
      \hline
        100 & 39.9  & 66.5 & 2.0 & 19.8 \\
        200 & 79.0 & 131.6 & 3.9 & 39.2 \\
        500 & 211.3 & 352.2 & 9.7 & 97.2 \\
      \hline
\end{tabular}
\caption{Computation time for the full model and pairwise model as a function of the number of trials from MCMC approach for setting 2.}
\label{tab:computation_time_MCMC_s2}
\end{table}

\FloatBarrier
\clearpage

\section{Parameter recovery and simulation-based calibration of NPEs}
\label{ap:recovery}
\begin{figure}[!ht]
    \centering
    \includegraphics[width=1\linewidth]{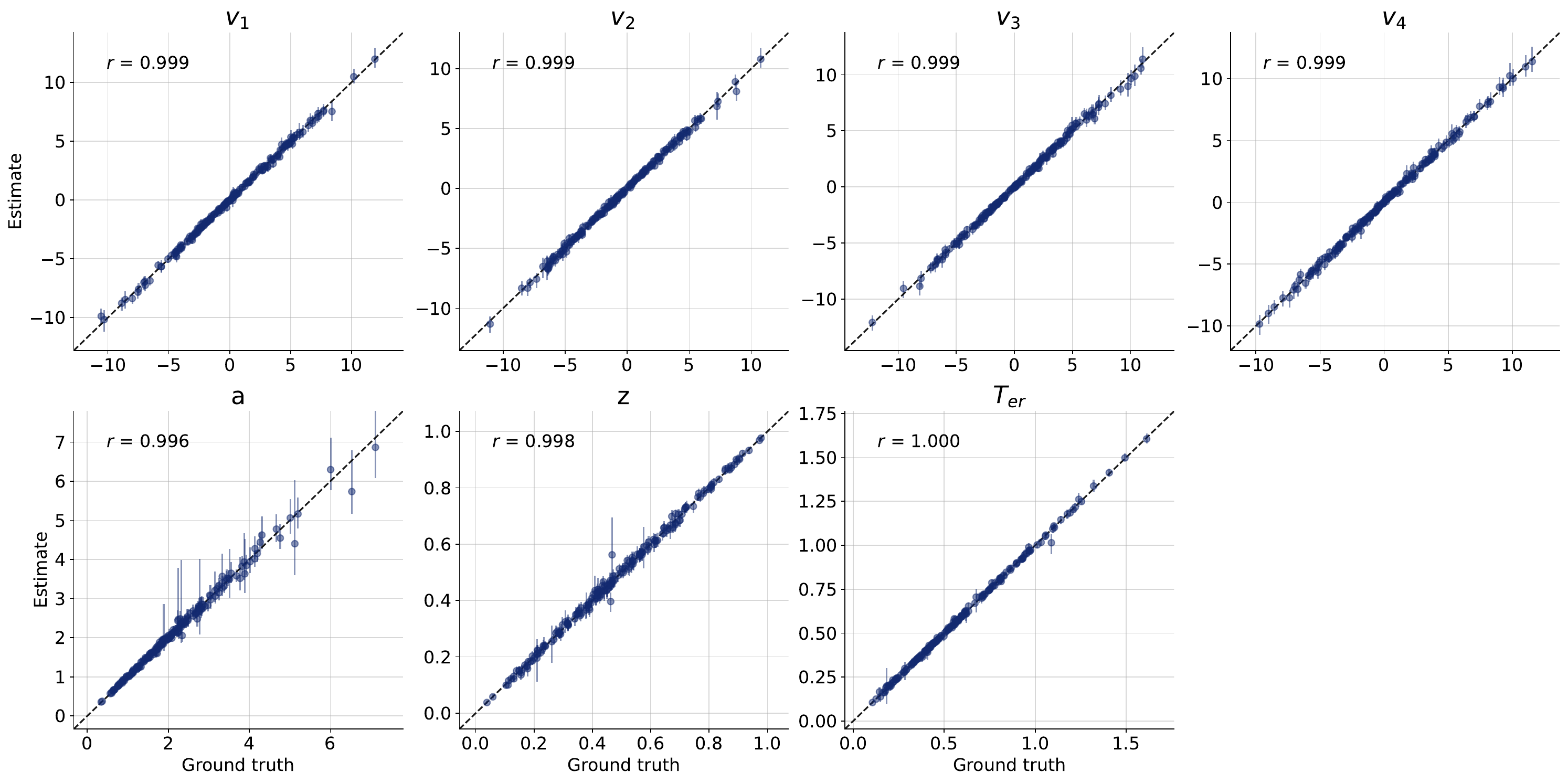}
    \label{fig:ABI_recovery_s1}
\caption{The parameter recovery of the full-model NPE in setting 1. The x-axis shows ground truth parameter values, the y-axis shows posterior mean estimates, the dashed line indicates perfect recovery, and the top-left value is the correlation between true and estimated values. }
\end{figure}

\begin{figure}[!ht]
    \centering
    \includegraphics[width=1\linewidth]{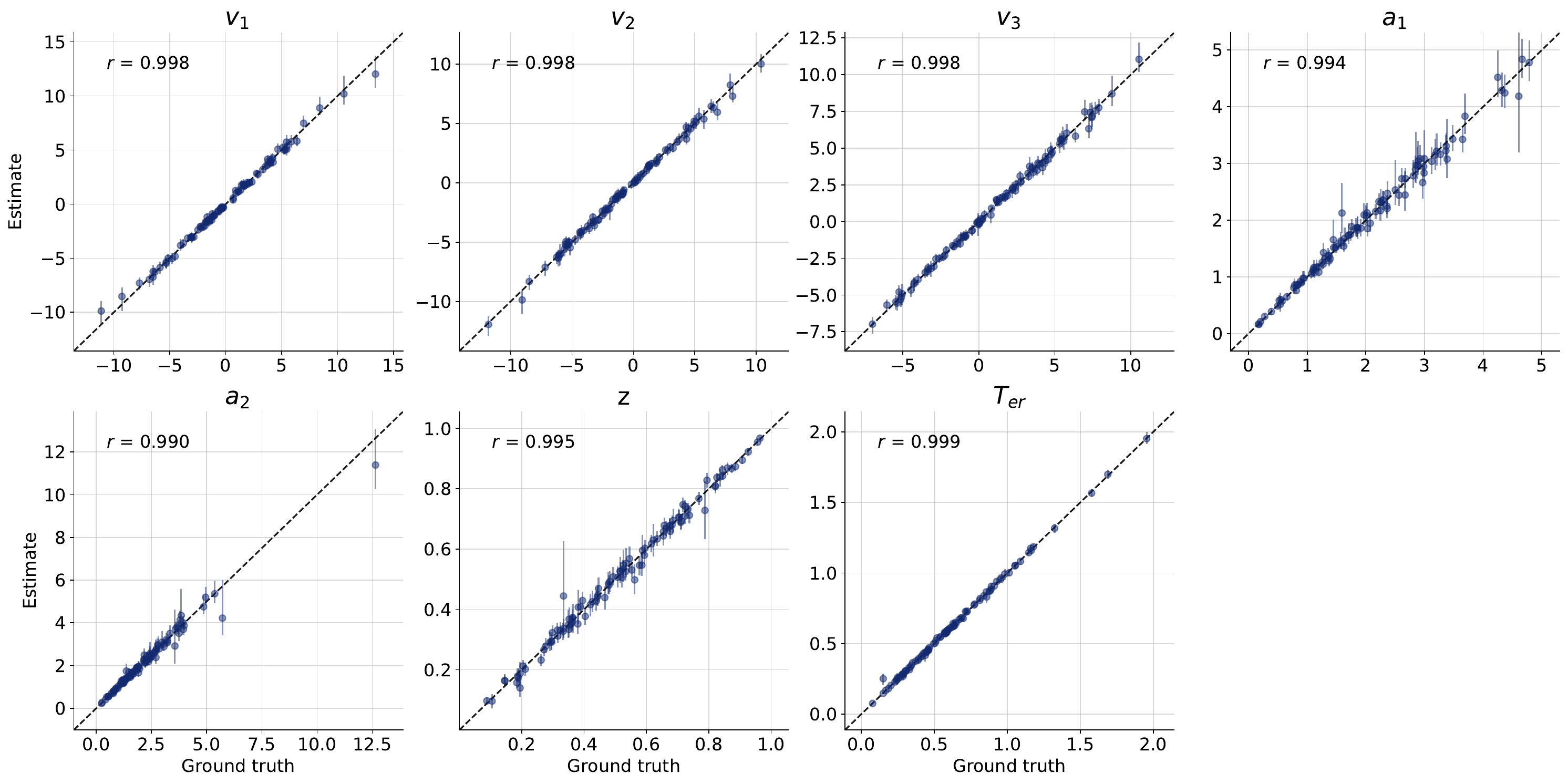}
    \caption{The parameter recovery of the full-model NPE in setting 2. The x-axis shows ground truth parameter values, the y-axis shows posterior mean estimates, the dashed line indicates perfect recovery, and the top-left value is the correlation between true and estimated values.}
    \label{fig:ABI_recovery_s2}
\end{figure}

\begin{figure}[!ht]
    \centering
    \includegraphics[width=1\linewidth]{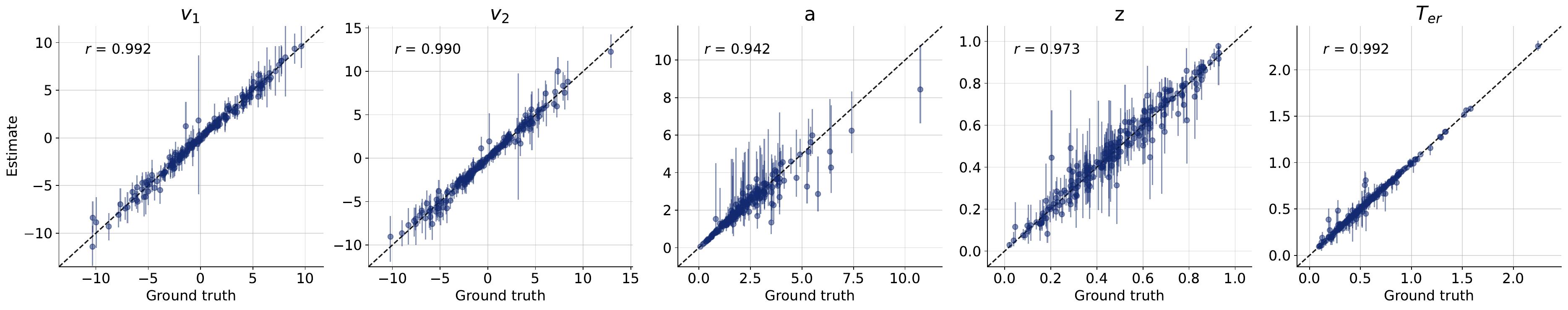}
    \caption{The parameter recovery of the pairwise NPE used in both settings. The x-axis shows ground truth parameter values, the y-axis shows posterior mean estimates, the dashed line indicates perfect recovery, and the top-left value is the correlation between true and estimated values.}
    \label{fig:ABI_recovery_pairwise}
\end{figure}

\begin{figure}[!ht]
    \centering
    \includegraphics[width=1\linewidth]{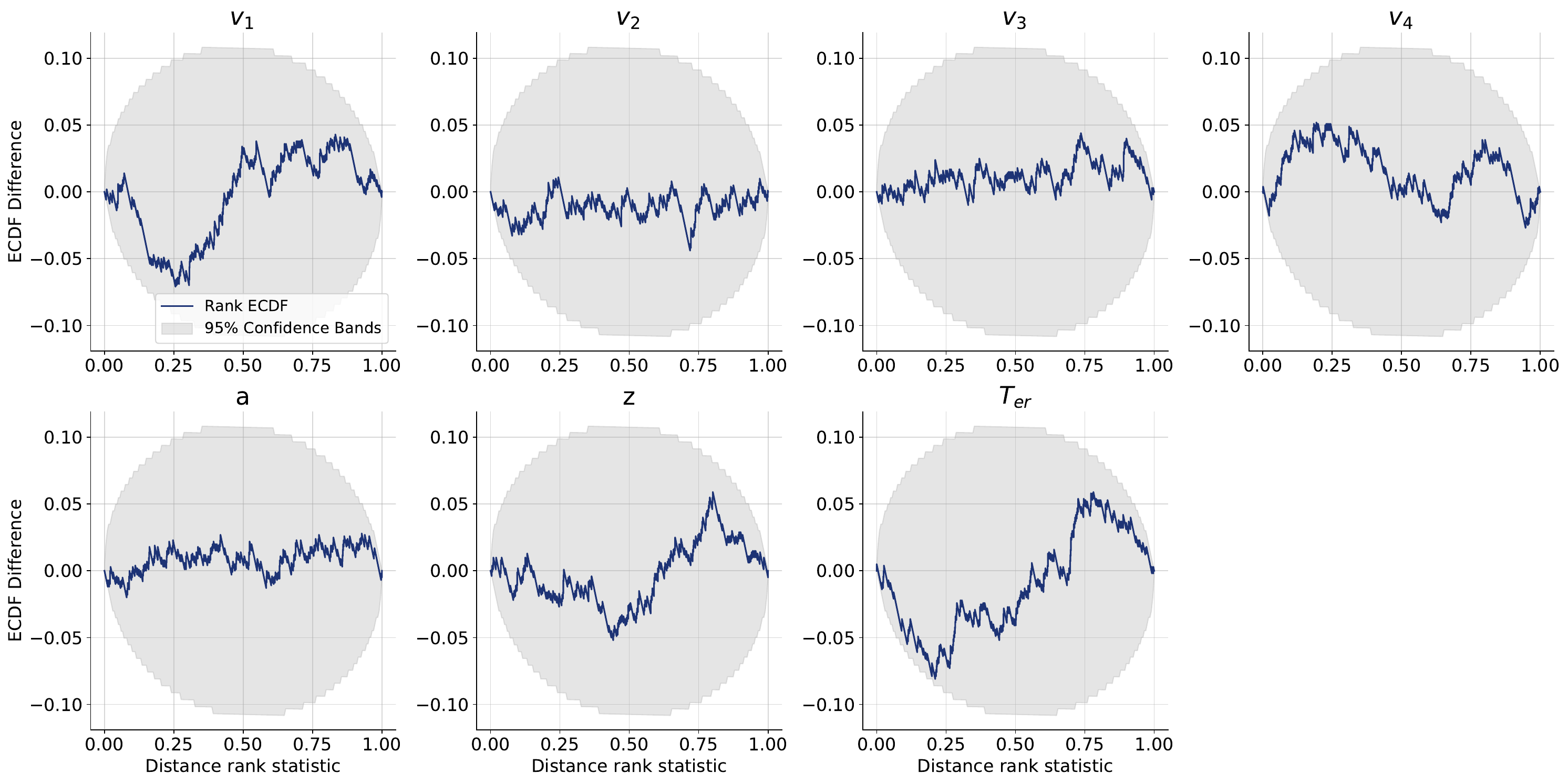}
    \label{fig:ABI_sbc_s1}
\caption{The empirical cumulative distribution function (ECDF) of rank statistics plotted against the uniform ECDF for the full-model NPE in setting 1. If the posterior ranks of the prior draws distributed uniformly, the estimator yielded correct posteriors. For details of rank statistics, see  Säilynoja et al. (\citeyear{sailynoja2022graphical}).}
\end{figure}

\begin{figure}[!ht]
    \centering
    \includegraphics[width=1\linewidth]{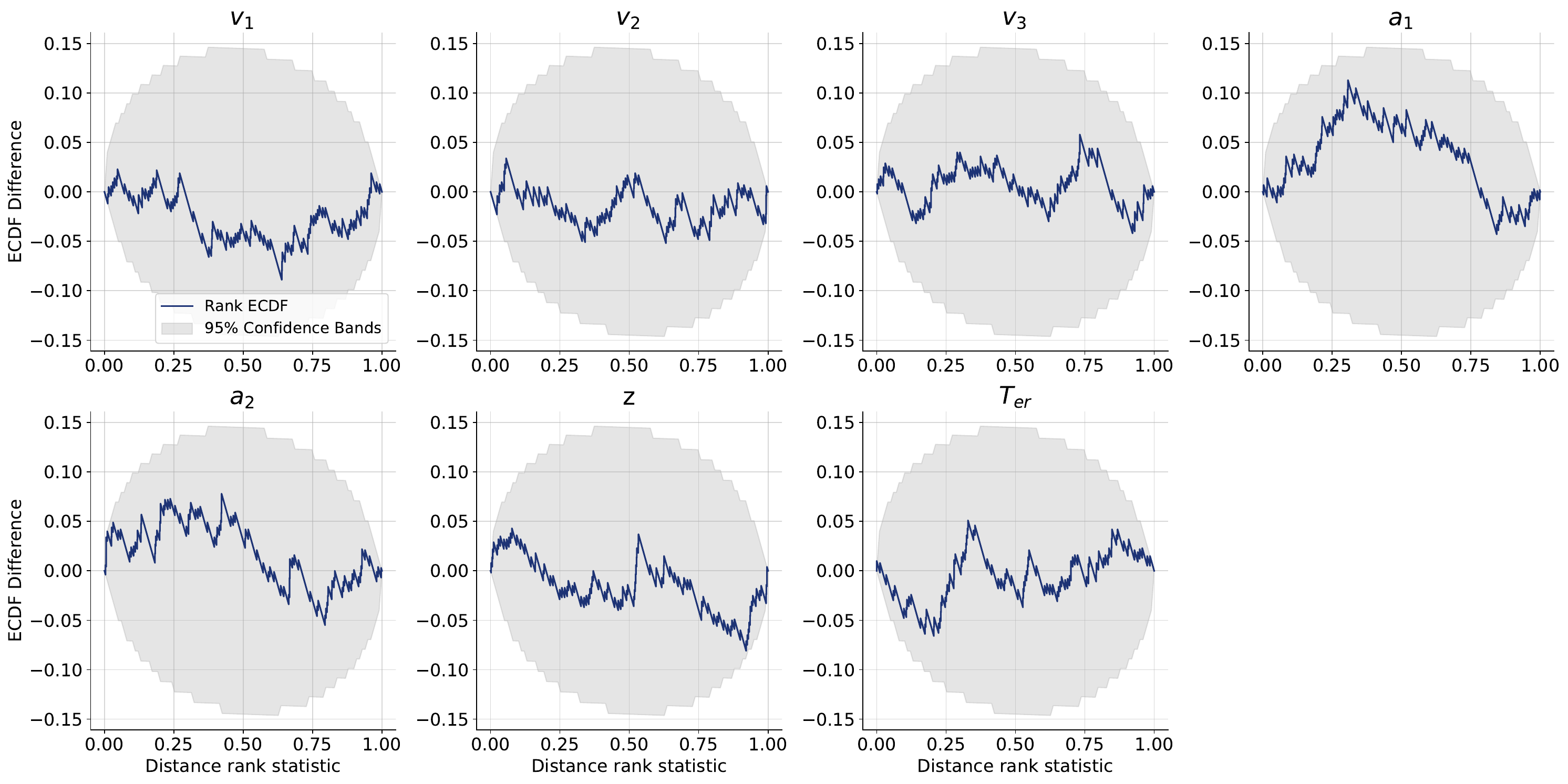}
    \label{fig:ABI_sbc_s2}
\caption{The empirical cumulative distribution function (ECDF) of rank statistics plotted against the uniform ECDF for the full-model NPE in setting 2. If the posterior ranks of the prior draws distributed uniformly, the estimator yielded correct posteriors. For details of rank statistics, see  Säilynoja et al. (\citeyear{sailynoja2022graphical}).}
\end{figure}

\begin{figure}[!ht]
    \centering
    \includegraphics[width=1\linewidth]{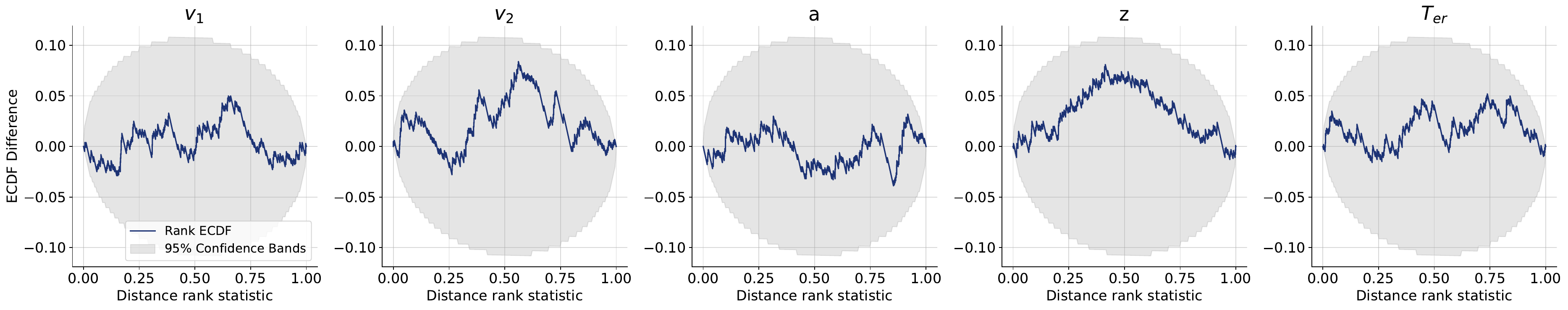}
    \label{fig:ABI_sbc_pairwise}
\caption{The empirical cumulative distribution function (ECDF) of rank statistics plotted against the uniform ECDF for the pairwise NPE used in both settings. If the posterior ranks of the prior draws distributed uniformly, the estimator yielded correct posteriors. For details of rank statistics, see  Säilynoja et al. (\citeyear{sailynoja2022graphical}).}
\end{figure}

\clearpage
\section{Full model results for real data example dataset}
\label{app: full_model_realexample}
\begin{figure}[!ht]
    \centering
    \includegraphics[width=1\linewidth]{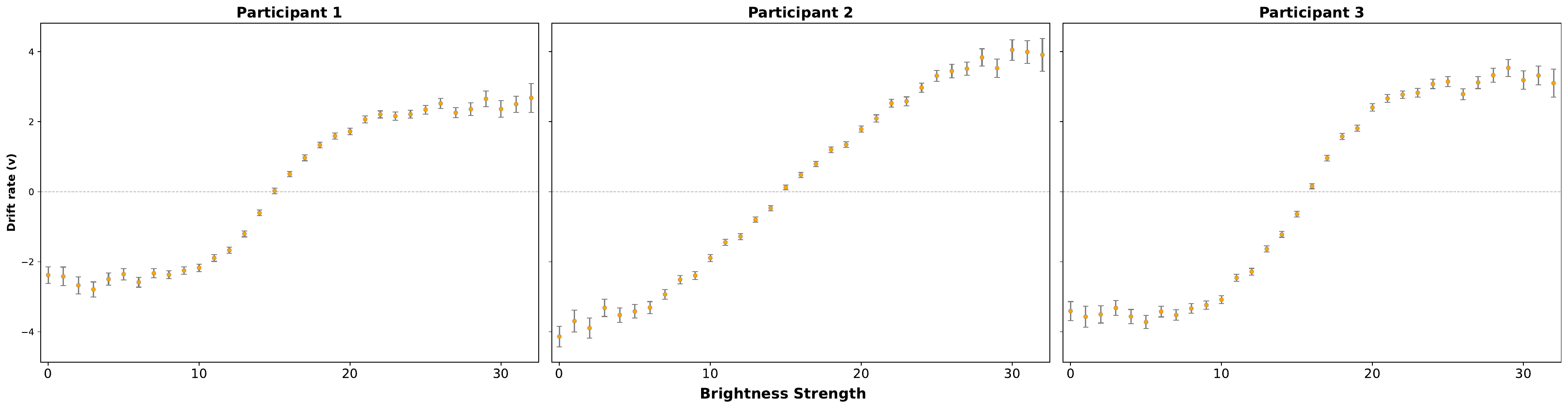}
    \caption{Estimated drift rates across 33 brightness strength levels from the full MCMC model. Error bars denote posterior means $\pm$ standard deviation.}
    \label{fig:Full_MCMC_real.data.example}
\end{figure}

\begin{table}[!ht]
\centering
\begin{tabular}{lcccc}
\hline
\textbf{Participant} & \textbf{$a_{\text{speed}}$} & \textbf{$a_{\text{accuracy}}$} & \textbf{$z$} & \textbf{$T_{er}$} \\
\hline
1 & 0.829(0.007)& 2.092(0.019)& 0.477(0.004)& 0.196(0.000) \\
2 & 0.791(0.007)& 2.051(0.017)& 0.512(0.004)& 0.195(0.000)\\
3 & 1.079(0.010)& 1.852(0.015)& 0.517(0.004)& 0.195(0.000)\\
\hline
\end{tabular}
\caption{Parameter estimates from the full MCMC model by participant}
\label{tab:realdata_full_mcmc}
\end{table}

\end{appendices}

\end{document}